\documentclass[10pt,twocolumn,letterpaper]{article}

\pdfoutput=1
\usepackage{iccv}
\usepackage{times}
\usepackage{epsfig}
\usepackage{graphicx}
\usepackage{amsmath}
\usepackage{amssymb}
\usepackage{booktabs}
\usepackage{float}
\usepackage{subcaption}
\usepackage{soul}
\usepackage{stfloats}
\usepackage{xcolor,colortbl}
\usepackage{mathtools}
\usepackage{authblk}
\usepackage[toc,page,title]{appendix}
\usepackage{multirow}
\usepackage{listings}

\newfloat{listing}{tb}{lst}{}
\floatname{listing}{Listing}

\usepackage[pagebackref,breaklinks,colorlinks]{hyperref}
\usepackage[capitalize]{cleveref}
\crefname{section}{Sec.}{Secs.}
\Crefname{section}{Section}{Sections}
\Crefname{table}{Table}{Tables}
\crefname{table}{Tab.}{Tabs.}

\newcommand{\red}{\color{red}}
\newcommand{\blue}{\color{blue}}
\newcommand{\black}{\color{black}}
\def\@fnsymbol#1{\ensuremath{\ifcase#1\or *\or \dagger\or \ddagger\or
   \mathsection\or \mathparagraph\or \|\or **\or \dagger\dagger
   \or \ddagger\ddagger \else\@ctrerr\fi}}
\newcommand{\ssymbol}[1]{^{\@fnsymbol{#1}}}

\definecolor{Gray}{gray}{0.85}
\definecolor{LightCyan}{rgb}{0.88,1,1}

\newcolumntype{a}{>{\columncolor{Gray}}c}
\newcolumntype{b}{>{\columncolor{white}}c}

\iccvfinalcopy % *** Uncomment this line for the final submission

\def\iccvPaperID{9874} % *** Enter the ICCV Paper ID here

\ificcvfinal\fi

\begin{document}

%%%%%%%%% TITLE
\title{A Generative Model for Digital Camera Noise Synthesis}

\author[1, 2]{Mingyang Song}
\author[2]{Yang Zhang}
\author[2]{Tun\c{c} O. Ayd{\i}n}
\author[1, 2]{Elham Amin Mansour}
\author[2]{Christopher Schroers}

\affil[1]{ETH Zurich, Switzerland}
\affil[2]{DisneyResearch$|$Studios, Switzerland}
\affil[ ]{\tt {misong@student.ethz.ch, yang.zhang@disneyresearch.com}}

\maketitle
\ificcvfinal\thispagestyle{empty}\fi

\begin{abstract}
Noise synthesis is a challenging low-level vision task aiming to generate realistic noise given a clean image along with the camera settings. To this end, we propose an effective generative model which utilizes clean features as guidance followed by noise injections into the network. Specifically, our generator follows a UNet-like structure with skip connections but without downsampling and upsampling layers. Firstly, we extract deep features from a clean image as the guidance and concatenate a Gaussian noise map to the transition point between the encoder and decoder as the noise source. Secondly, we propose noise synthesis blocks in the decoder in each of which we inject Gaussian noise to model the noise characteristics. Thirdly, we propose to utilize an additional Style Loss and demonstrate that this allows better noise characteristics supervision in the generator. Through a number of new experiments, we evaluate the temporal variance and the spatial correlation of the generated noise which we hope can provide meaningful insights for future works. Finally, we show that our proposed approach outperforms existing methods for synthesizing camera noise.

\end{abstract}

\begin{figure}
    \centering
    \includegraphics[width=0.9\linewidth]{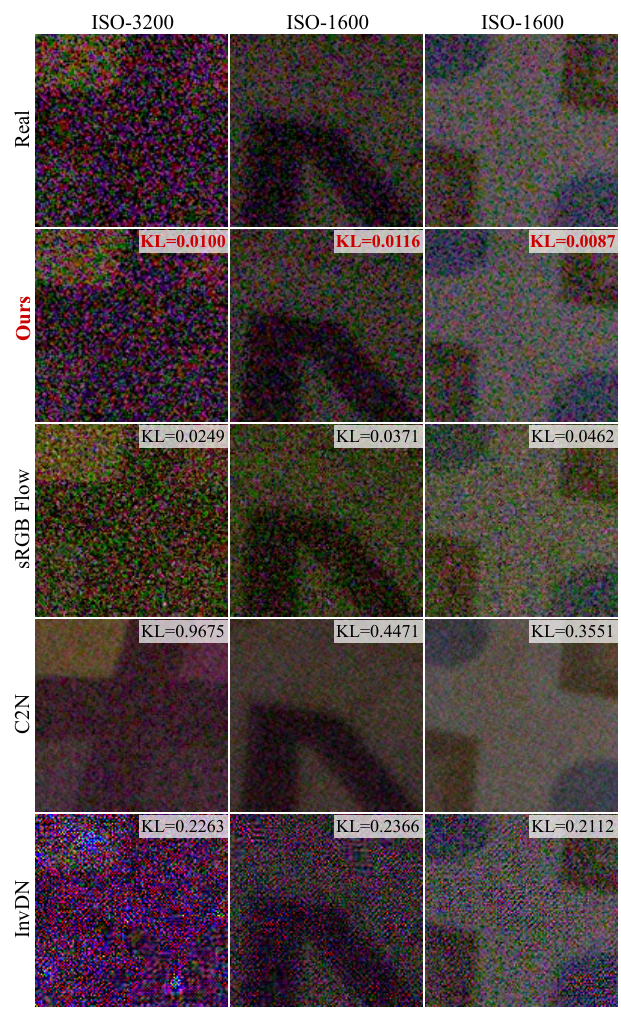}
    \caption{Comparison of real and deep-learning-based synthetic noisy images.}
    \label{fig:tesaerl}
\end{figure}

%%%%%%%%% BODY TEXT
\section{Introduction}
\label{sec:intro}
Digital camera noise modeling aims to simulate the true distribution of sensor noise conditioned on the captured photon densities and camera settings. Many practical image and video processing applications can benefit from an accurate noise synthesis model. For instance, synthetic noisy images can significantly enrich training datasets for image and video denoising. Due to the difficulty of obtaining clean-noisy image pairs, many denoising methods have utilized Additive White Gaussian Noise (AWGN)\cite{tassano2020fastdvdnet, liang2021swinir}. However, models trained on AWGN cannot be applied directly to real camera denoising tasks because the domain gap constrains the models' performance on real data. On the other hand, in modern digital movie production noise is not necessarily a degradation, but rather an artistic ingredient demanded by the artists and consumers as in the case of film grain\cite{ameur2022deep}. However, due to its random nature and high entropy, noise is inherently difficult to compress. To reduce the bandwidth requirements of noisy content, transmitting the denoised content and finally adding the noise back at the client is attracting more and more attention in modern video codecs \cite{norkin2018film}. In this sense, generating realistic noise with stochastic variation in spatial and temporal dimensions is a crucial, albeit challenging task to satisfy both aforementioned use cases.

 The task of modeling camera noise can be split into two categories according to the color space of the content. The first one is generating noise before the Image Signal Processor (ISP), namely in rawRGB color space\cite{maleky2022noise2noiseflow, abdelhamed2019noise, wei2020physics, zou2022estimating, zhang2021rethinking}. Another category is modeling the noise after such an image processing pipeline, namely in sRGB color space\cite{kousha2022modeling, jang2021c2n}. The ISP introduces non-linear transformations and spatial mixing to the raw data. Therefore the noise distributions in these two color spaces deviate from each other. Most works only focus on one of the color spaces. However, our model can generalize to both cases.

The essence of noise modeling can be interpreted as mapping a simple parametric distribution to a more complex one, which can be achieved through generative modeling, \eg through the use of a Generative Adversarial Network (GAN)\cite{goodfellow2020generative}. However, when the dataset size is limited, the adversarial loss cannot perfectly measure the distance between two distributions, which makes the training slow and unstable. To solve this problem, some works\cite{jang2021c2n, cai2021learning, monakhova2022dancing, yue2020dual} focused on designing specialized adversarial loss functions and additional supervision on the generator. Another popular generative model, Normalizing flow~\cite{rezende2015variational}, is also exhaustively explored in the noise modeling task. The approaches proposed in \cite{maleky2022noise2noiseflow, abdelhamed2019noise, kousha2022modeling} use invertible neural networks to map camera noise to Gaussian noise and run the inverted model to sample from a Gaussian distribution.

In this paper, we propose a simple but effective generator based on a conditional GAN model. Given different conditions (\eg camera settings, sensor types), our model can generate different types of camera noise from various distributions accurately. To stabilize the training we include a Style Loss\cite{gatys2016image}, which has been originally designed for measuring the distance between two images in style space for the task of style transfer. We propose to utilize the Style Loss as an additional supervision for the Generator as a measure of distribution similarity between real and synthesized noise samples. Qualitative and quantitative evaluations indicate that our method outperforms all the other recent works in the sRGB space. Meanwhile, our method achieves comparable performance in rawRGB space as networks that have specifically been designed for operating in that space. Moreover, we design new experiments on the Smartphone Image Denoising Dataset (SIDD)\cite{abdelhamed2018high}, which further verify that the distribution of our generated noise is close to the ground truth from spatial and temporal perspectives respectively.

\section{Related Work}
\label{sec:related_works}
\textbf{Traditional Methods.} Traditional methods model the noise on each pixel using simple distributions, such as additive white Gaussian noise (AWGN). 
%To increase the complexity, 
The heteroscedastic Gaussian model also takes into account the dependency on pixel intensity by splitting the variance into two parts according to their signal dependency: 
%splits the noises into two parts, which can be described as:
\begin{equation}
    n\sim\mathcal{N}(0,\sigma_{dep}^2(I)+\sigma_{ind}^2),
    \label{eq:g_gaussian}
\end{equation}
where $I$ is the intensity of the clean pixel value, $\sigma_{dep}^2(I)$ and $\sigma_{ind}^2$ represent the variance of signal-dependent and independent noises, respectively. 
%Since such methods can only generate fine-grained noise, they are often used for synthesizing noises in rawRGB space. 

\textbf{Physics-based Methods.} Wei \etal~\cite{wei2020physics} physically modeled different components of the noises (\eg shot noise, read noise, etc.) and their relationship with overall system gain through the linear least-square fitting. Following this work, Zou \etal~\cite{zou2022estimating} used contrastive learning to estimate the camera parameters from noisy images to save the labor of manually calibrating the noises.

\textbf{Generative Adversarial Network.} Jang \etal~\cite{jang2021c2n} explicitly sampled the content-dependent and independent noise in deep feature space and trained the model with unpaired data. To ensure the correctness of synthetic noises' pattern and magnitude without sacrificing randomness, Jang \etal~\cite{jang2021c2n} added a penalty to the mean value of generated noises. Cai \etal~\cite{cai2021learning} used the perceptual loss and aligned the synthetic and real noises on the image domain with an additional denoiser. Fu \etal~\cite{fu2023srgb} used a lightweight model to predict the global noise level from noisy images and a mean filter to approximate the local level map.

\textbf{Invertible Network.} Abdelhamed \etal~\cite{abdelhamed2019noise} used normalizing flow, that conditions on sensor gain and camera brandmark, to map the camera noise to a base measure (i.e., Gaussian noise), and sample noises through the inverted network in rawRGB space. Based on it, Kousha \etal~\cite{kousha2022modeling} improved the model and applied it in the sRGB space. Maleky \etal~\cite{maleky2022noise2noiseflow} trained a self-supervised denoiser~\cite{lehtinen2018noise2noise} and Noise Flow~\cite{abdelhamed2019noise} model jointly without clean images, and achieved promising performance in both denoising and noise synthesizing tasks. Liu \etal~\cite{liu2021invertible} used an invertible network to separate clean image signal and noise signal in latent space and performed the denoising task simultaneously.

\section{Method}
\label{sec:method}
\begin{figure*}
    \centering
    \subcaptionbox{Conditional generator\label{fig:cfg_nin}}{%
        \includegraphics[height=5.5cm]{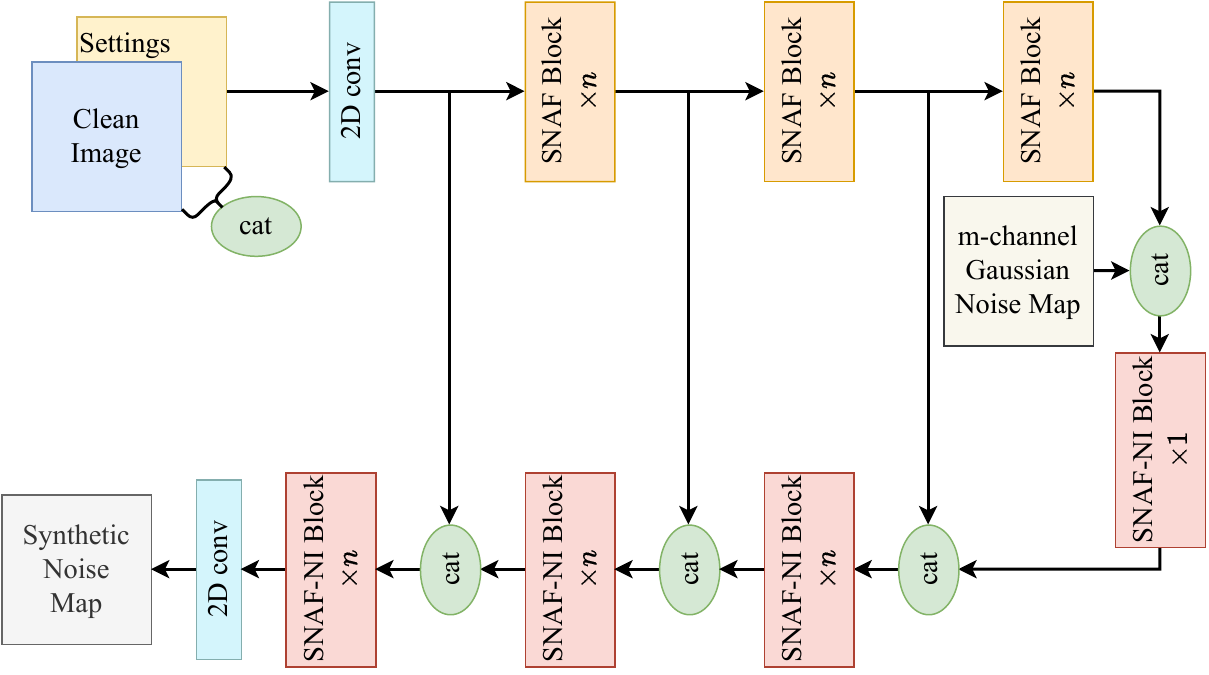}%
    }
    \enskip
    \subcaptionbox{Conditional discriminator\label{fig:disc}}{%
        \includegraphics[height=5.5cm]{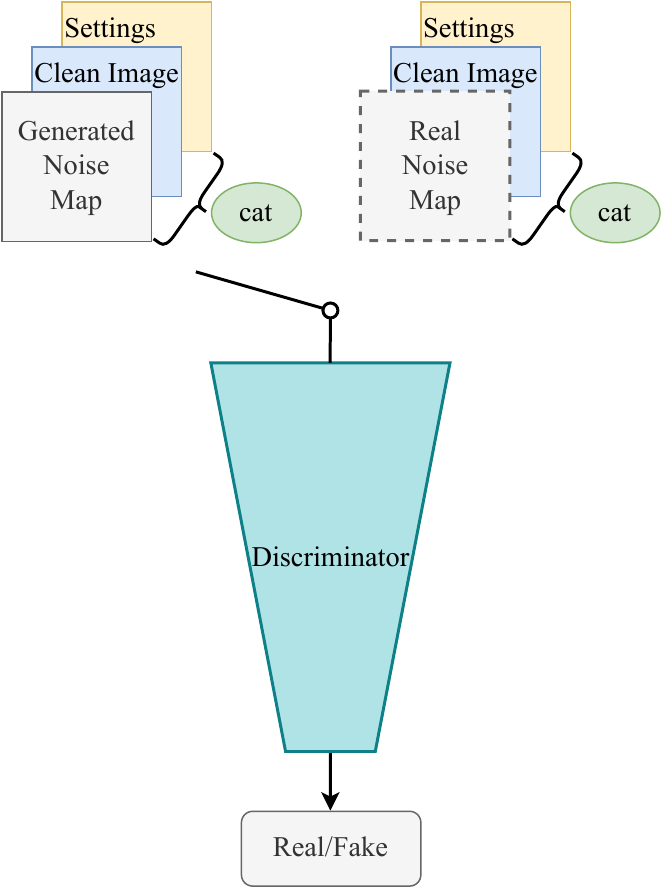}%
    }
    \caption{The architecture of our generator and discriminator.}
    \label{fig:overal_arch}
\end{figure*}

Our model is trained on clean-noisy image pairs. It predicts the residuals between the clean image and the corresponding noisy image, which we call the noise map. 
%We realized that the separated noise signal and the clean image signal simplified the work for the discriminator to judge whether a sample belongs to the real distribution, which further stabilized the GAN training. 
This choice was motivated from our observation that residual prediction resulted in more stability during GAN training. In order to train a conditional GAN for artistic control we feed the additional control information (camera brandmark, ISO, shutter speed, etc.) to the generator besides the clean image. We introduce the concept of noise injection into our generator to imitate the stochastic variation of real noises which is added onto the concept of StyleGAN2\cite{karras2020analyzing}. The process of generating one noise map can be formulated as:
\begin{equation}
  \hat{n} = G(n, I, CM),
  \label{eq:1}
\end{equation}
where $\hat{n}$ is the synthetic noise map, $G$ is the generator, $I$ is the clean image, $CM$ is the control information, which we call control map (CM) throughout the rest of the paper, and $n$ is the Gaussian noise with a standard deviation of one. On the other hand, we feed the real or synthetic noise map, the corresponding clean image, and control information to the discriminator (Fig.~\ref{fig:disc}), which can be described as:
%. The functionality of the discriminator can be seen in Fig.~\ref{fig:disc} and described as:
\begin{equation}
  p = D(n^{*}, I, CM),
  \label{eq:2}
\end{equation}
where $n^{*}$ represents either the real ($\tilde{n}$) or synthetic ($\hat{n}$) noise map, $D$ is the discriminator and $p$ is the discriminator score. To improve readability, in the rest of the paper we will use a shorthand notation $D(\cdot) \coloneqq D(\cdot, I, CM)$ and $G(\cdot) \coloneqq G(\cdot, I, CM)$ to refer to the discriminator and the generator.

Fig. \ref{fig:cfg_nin} shows the architecture of our generator. 
We start by extracting features from the clean input image and control map through a series of encoder blocks. We then concatenate one $m$-channel i.i.d. standard Normal noise map to the transition point between the encoder and decoder. This serves as the initial seed for the synthetic noise that our model produces. The computation then proceeds with a series of decoder blocks with noise injections (described later in Sec.~\ref{subsec:SNAF-NI}) that gradually convert the initial noise map to the desired camera noise distribution.
%Firstly, we use encoders to extract clean features from the clean image with the control map. Secondly, we concatenate one m-channel Gaussian noise map to the latent bottleneck, which acts as the source of the synthetic noise. The noises are sampled independently on each 3D position with a standard deviation of one. Next, we use decoders with noise injections inside (which will be described in Section~\ref{subsec:SNAF-NI} in detail) to convert the Gaussian noise map into synthetic camera noise. 
To reinforce the content dependency, we condition the synthesis process on the clean features extracted in each stage, i.e., concatenating the clean features to the noise generation layers with skip connections. We use concatenation instead of addition for better separation of the synthesized noise and the clean features. 
%The spatial resolution of all features keeps the same as the input because we observed severe repetitive artifacts introduced by up/down samplings.
Throughout our model the image resolution remains unchanged, as we empirically observed that up- and down-sampling may produce visible patterns in the final prediction. As our conditional discriminator we use the same architecture as in ~\cite{karras2020analyzing}.

\subsection{SNAF Block and SNAF-NI Block} 
\label{subsec:SNAF-NI}
\begin{figure}[]
    \centering
    \subcaptionbox{SNAF Block.\label{fig:snaf}}{%
        \includegraphics[height=7cm]{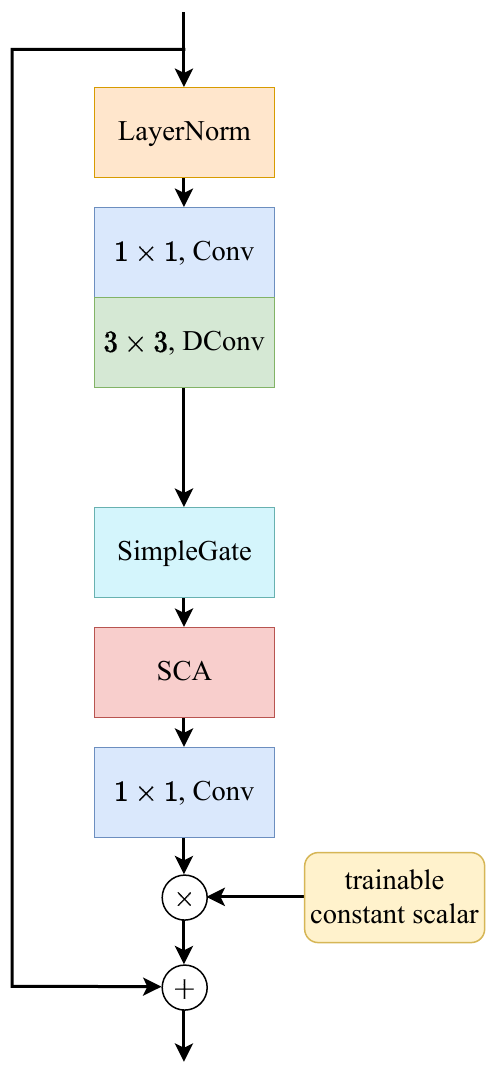}%
    }
    \enskip
    \subcaptionbox{SNAF-NI Block.\label{fig:snaf-ni}}{%
        \includegraphics[height=7cm]{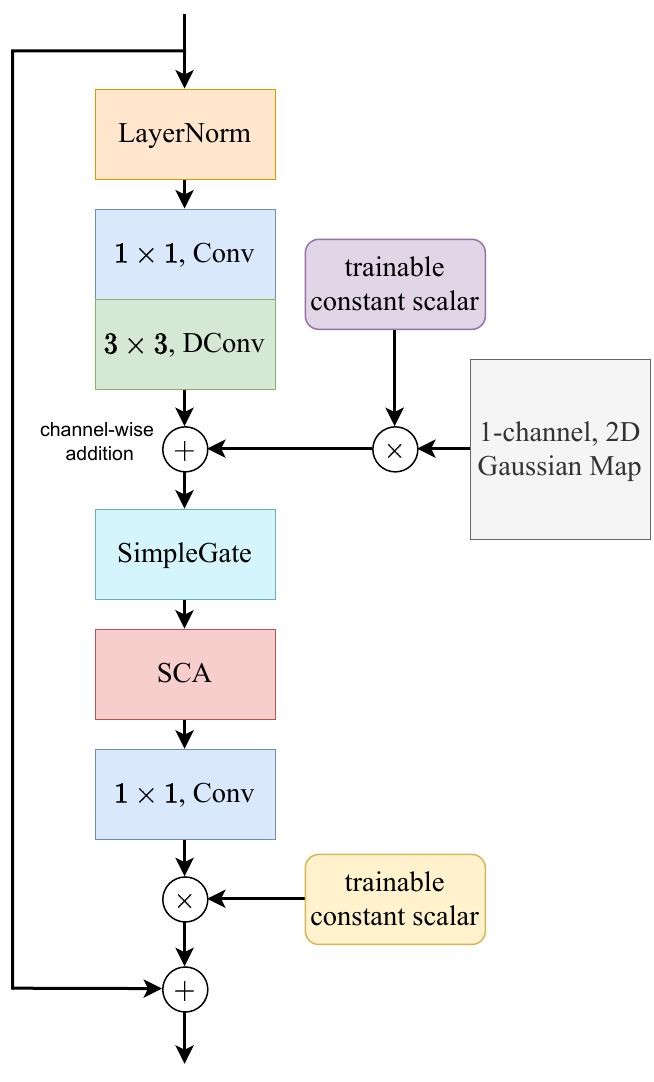}%
    }
    \caption{Components of SNAF and SNAF-NI block.}
    \label{fig:blovk}
\end{figure}

Each encoder block used in our generator contains a sequence of modified Nonlinear Activation Free (NAF)~\cite{chen2022simple} blocks. The original NAF block consists of two sequences of layers with skip connections around them. The first sequence of layers contains all important ingredients, namely Layer Normalization~\cite{ba2016layer}, a gating mechanism~\cite{shazeer2020glu}, and channel attention~\cite{hu2018squeeze}. We found that the second sequence of layers in a NAF block is not needed for obtaining high quality results in our task. In order to end up with a simpler and leaner architecture, we therefore propose to only take the first sequence of layers in a NAF block and refer to this as Simple NAF (SNAF) block shown in Fig.~\ref{fig:snaf}. 

For our decoder, we propose a SNAF block with noise injection (SNAF-NI). We place the noise injection between the convolution layer and the simple gating layer as shown in Fig.~\ref{fig:snaf-ni}, which is added onto the concept of \cite{karras2020analyzing}. The injected noise is a one-channel Gaussian noise map with the same spatial resolution as the deep features. The noise value is sampled independently on each spatial position. We use a trainable scalar to control the variance of this Gaussian noise map and add it to each channel of the features. We then use another trainable scalar to control the contribution of the noise injected features.

\subsection{Loss Functions}
\textbf{GAN Loss.} We use the standard conditional GAN loss as discriminator loss and adversarial loss, which can be described as:
\begin{equation}
    \begin{array}{l}
        \underset{G}{\min} \:\underset{D}{\max} \:V(D,G)= \\
        \underset{\tilde{n},I,CM\sim P_{data}}{\mathbb{E}}\{\log D(\tilde{n}) + 
        \underset{n\sim P_{g}(n)}{\mathbb{E}}[\log(1-D(G(n)))]\},
    \end{array}
    \label{eq:gan_loss}
\end{equation}
where $P_{data}(\tilde{n},I,CM)$ represents sampling one clean-noise image pair and its corresponding settings, and $P_g(n)$ represents sampling the Gaussian noise maps in the transition point and all decoder blocks. More specifically, we use the Squared Error as the loss on the discriminator scores, which can be described as:
\begin{equation}
    \mathcal{L}_{disc}=[(1-D(\tilde{n}))^2+D(\hat{n})^2]/2,
    \label{eq:disc_loss}
\end{equation}
\begin{equation}
    \mathcal{L}_{adv}=(1-D(\hat{n}))^2.
    \label{eq:adv_loss}
\end{equation}

\textbf{Style Loss.} The style loss has initially been used in Style Transfer~\cite{gatys2016image} to measure the distance of two images in style space. Commonly, the style loss is defined as the distance between two features' Gram Matrices. \cite{li2017demystifying} pointed out that minimizing the distance between Gram Matrices can be interpreted as Maximum Mean Discrepancy (MMD) minimization process with the second-order polynomial kernel, which measures the distance between two distributions through the samples:
\begin{equation}
    \mathcal{L}_{style}=\frac{1}{4M^2}\text{MMD}^2[\hat{n},\tilde{n}],
    \label{eq:mmd}
\end{equation}
where $M$ is the number of pixels in one patch, $\hat{n}$ and $\tilde{n}$ represent the synthesized and the real noises respectively. As such the style loss effectively reasons on statistical properties which makes it a promising candidate to supervise the noise generation process. We use the pre-trained VGG-Network~\cite{simonyan2014very} to extract the perceptual features of the real and synthetic noise map and compute the average Style Loss with uniform weights over different layers, which can be formulated as:
\begin{equation}
    \mathcal{L}_{style}=\frac{1}{L}\sum_{l=1}^{L}\text{MSE}(Gram_{\tilde{n}}^{l},Gram_{\hat{n}}^{l}),
    \label{eq:style_loss}
\end{equation}
where $L$ is the number of layers, $L=5$ in this work, $\text{MSE}(\cdot,\cdot)$ is the Mean Squared Error loss, and $Gram^l$ is the Gram Matrix (GM) of the feature in layer $l$.

\section{Results}
\label{sec:results}
\subsection{Experimental Setup}
\label{subsec:experiments_setup}
\textbf{Dataset.} SIDD-Medium~\cite{abdelhamed2018high} consists of 10 scenes and 138 settings, and their combinations result in 160 instances. For each instance, there are two pairs of clean-noisy images. 

\textbf{\label{EM1}Evaluation method-1 (EM-1): }\cite{abdelhamed2019noise, kousha2022modeling, maleky2022noise2noiseflow} removed images with rare ISO levels in the dataset and provided lists that specified the training and testing set in the codes\footnote{\label{note1}https://github.com/SamsungLabs/Noise2NoiseFlow}. We used the same number of condition information (i.e., camera brandmark and ISO level) for our control map and trained and tested our networks on the same lists\textsuperscript{\ref{note1}} for fair comparison (as shown in Table~\ref{table:test_srgb}). 

\textbf{\label{EM2}EM-2: }We observed the imbalance of camera brandmarks and settings in the provided lists\textsuperscript{\ref{note1}}. To fully utilize the diversity of SIDD~\cite{abdelhamed2018high} and keep consistent with ~\cite{jang2021c2n, liu2021invertible}, we cropped each scene instance into several $512\times512$ patches and randomly select 80\% from each scene to form the training set, and the rest 20\% patches are used as the testing set. As a result, both the training and testing set cover all ISO levels and the imbalance is reduced. Besides the camera brandmark and ISO level, we add shutter speed, illuminant temperature, and brightness code to the control map for this training and testing settings (as shown in Table~\ref{table:test_srgb}).

\textbf{Metrics.} Due to the random nature of noise we use empirical statistical results to evaluate our methods. For EM-1, we follow the method described in ~\cite{abdelhamed2019noise, kousha2022modeling, maleky2022noise2noiseflow} by calculating the metrics on each patch with a spatial resolution of $32\times32$ and averaging the metrics on all testing patches. Following ~\cite{kousha2022modeling}, the histogram range in sRGB space is $[-260, 261]$, and the bin width is 4. Following ~\cite{maleky2022noise2noiseflow}, the histogram range in rawRGB space is $[-0.1,0.1]$, and the bin width is $0.2/64$. Such process can be described in a general form as:
\begin{equation}
    bin^{p}_{n^*}(i)=\frac{1}{CHW}\sum_{c,h,w}^{C,H,W}\mathbf{1}_{\{[l_i,r_i]\}}(n^*_{c,h,w}),
    \label{eq:bin}
\end{equation}
where $H,W=32$ represents the height and width of one patch, $C$ is the color channel ($C=3$ in sRGB space and $C=4$ in rawRGB space), $\mathbf{1}_{\{[l_i,r_i]\}}(n^*_{c,h,w})$ is the indicator function, and $bin^{p}_{n^*}(i)$ is the normalized value of $i$th bin of the patch $p$. Following ~\cite{kousha2022modeling, maleky2022noise2noiseflow}, we compute the forward Kullback-Leibler (KL) divergence to measure the distance between two histograms and the average of them on all patches in the testing set:
\begin{equation}
    D_{KL}^{f}(p)=\sum_{i}^{\#bins}bin^p_{\tilde{n}}(i)\log\frac{bin^p_{\tilde{n}}(i)}{bin^p_{\hat{n}}(i)},
    \label{eq:kl_per_patch}
\end{equation}
% \begin{equation}
%     D_{KL}^f=\frac{1}{\#patches}\sum_{p}^{\#patches}D^f_KL(p),
%     \label{eq:avg_kl}
% \end{equation}
where $D_{KL}^f(p)$ is the KL divergence of each patch $p$. The final result $D_{KL}^f$ is the average computed on all patches. 

The above metric considers the spatial correlation of noises and reduces the effect of error cancelation but suffers from the sensitivity to the patch spatial resolution. Consequently, we follow the KL divergence calculation method described in ~\cite{jang2021c2n} (EM-2), which stacks all testing patches as one image and then calculates the histogram (see Appendix~\ref{appendix:eval} for details).

We also provide KL divergence results for each R, G and B channel in  Appendix~\ref{appendix:quantitative}.

\subsection{Comparison to existing methods}
\label{subsec:quantitative_results}

\begin{figure}[]
    \centering
    \includegraphics[width=1.0\linewidth]{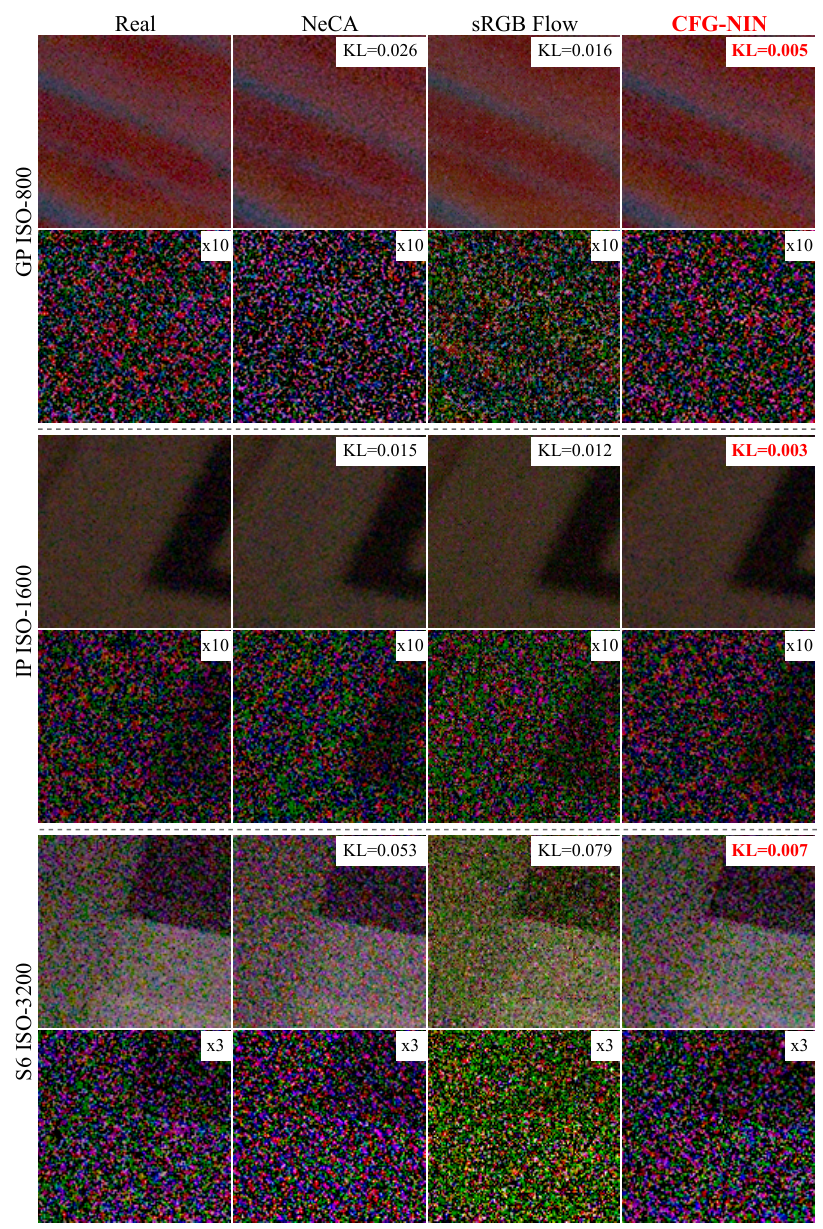}
    \caption{Visual comparison with the baseline methods on noise modeling task in sRGB space. Zoom in for better visualization.}
    \label{fig:comp_visual}
\end{figure}

\begin{figure}[]
    \centering
    \includegraphics[width=\linewidth]{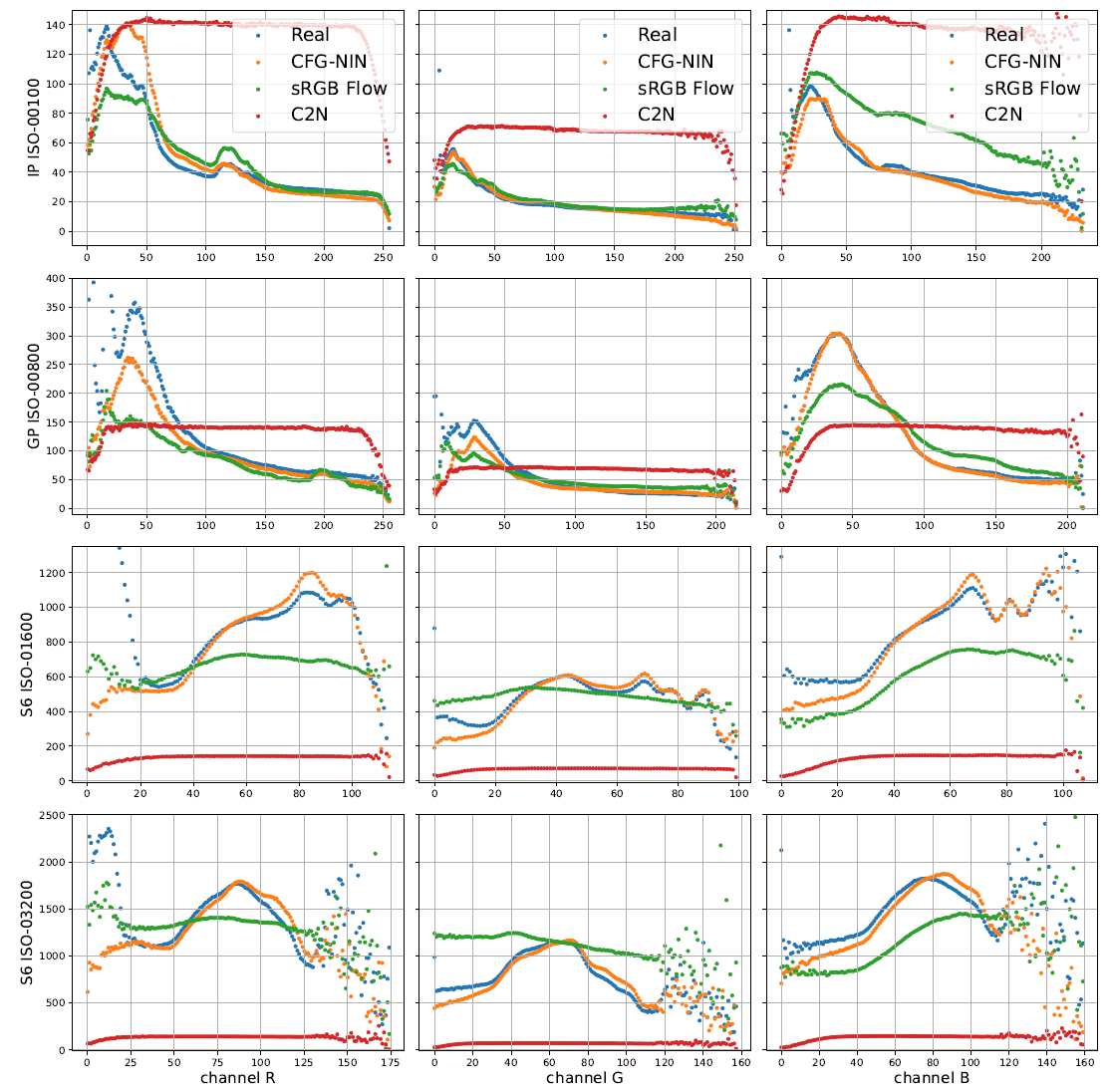}
    \caption{Clean image's intensity vs. variance of noise in sRGB space.
    All the sub-figures share the same legends as the first row for simplicity. The columns demonstrate the results on R, G and B channels separately.
    }
    \label{fig:var_vs_intensity}
\end{figure}

\textbf{sRGB noise modeling.} Table~\ref{table:test_srgb} shows the quantitative results of the noise modeling task in sRGB space by our network and six existing deep-learning-based methods. Under both evaluation methods, our method outperformed all the other deep-learning-based methods by a large margin. The visual comparison is shown in Fig.~\ref{fig:comp_visual}. Following ~\cite{kousha2022modeling}, we report the variance w.r.t the intensity in the sRGB space, which is shown in Fig.~\ref{fig:var_vs_intensity}. Given various control information, our model can synthesize noise with different noise distributions.

\begin{table*}[]
\small
\centering
\begin{tabular}{c|cccccc|cccccc|c}
\hline\hline
\multirow{2}{*}{Method} & \multicolumn{6}{c|}{$D_{KL}\downarrow$(EM-1, rare ISO levels excluded)} & \multicolumn{6}{c|}{$D_{KL}\downarrow$(EM-2$\times 10^2$, all ISO levels)} & \multicolumn{1}{c}{\multirow{2}{*}{\#Param}}\\ \cline{2-13}
 & G4 & GP & IP & N6 & S6 & Agg & G4 & GP & IP & N6 & S6 & Agg \\ \hline
InvDN\cite{liu2021invertible} & 0.111 & 0.008 & 0.077 & 0.051 & 0.161 &  0.092 & 3.360 & 1.793 & 5.483 & 5.806 & 5.261 & 5.204 & 2.64M\\
Noise Flow\cite{abdelhamed2019noise} & - & - & - & - & - & 0.198$^{\mathcal{x}}$ & - & - & - & - & - & - & 2.33K \\
sRGB Flow\cite{kousha2022modeling} & 0.044 & 0.059 & 0.020 & 0.062 & 0.050 & 0.044 & - & - & - & - & - & - & 6.62K \\ \hline
C2N\cite{jang2021c2n} & 0.217 & 0.426 & 0.114 & 0.230 & 0.541 & 0.335 & 8.970 & 18.54 & 3.070 & 38.69 & 45.29 & 14.06 & 2.15M \\ 
DANet\cite{yue2020dual} & 0.118 & 0.248 & 0.082 & 0.198 & 0.081 & 0.130 & 0.891 & 0.488 & 0.528 & 0.292 & 0.462 & 0.192 & 9.15M \\ 
NeCA\cite{fu2023srgb} & 0.074 & 0.030 & 0.027 & 0.081 & 0.046 & 0.044 & 1.043 & 0.469 & 0.472 & 1.027 & 0.842 & 0.300 & 8.09M \\ \hline
\textbf{CFG-NIN(ours)} & \textbf{0.026} & \textbf{0.015} & \textbf{0.012} & \textbf{0.037} & \textbf{0.025} & \textbf{0.021} & \textbf{0.071} & \textbf{0.067} & \textbf{0.106} & \textbf{0.047} & \textbf{0.151} & \textbf{0.055} & 1.19M\\ \hline\hline
\end{tabular}
\caption{Quantitative results of noise generating task in sRGB color space with EM-1 and EM-2 (the results are multiplied by $10^2$ for better readability.). The middle columns show the results w.r.t. camera brandmark, and the results in the last column (Agg) are aggregated on the whole testing set. The methods in the first three rows are invertible-network-based. The methods in the last four rows are GAN-based. ${\mathcal{x}}$: the results are taken from \cite{kousha2022modeling}. The lower the better.}
\label{table:test_srgb}
\end{table*}

\begin{table*}[]
\small
\centering
\begin{tabular}{ccccccc}
\hline\hline
Method     & G4 & GP & IP & N6 & S6 & Agg \\ \hline
Noise Flow~\cite{abdelhamed2019noise}$^{\mathcal{y}}$& - & 0.0180 & 0.0112 & - & 0.0469 &  0.0267 \\
N2N Flow~\cite{maleky2022noise2noiseflow}$^{\mathcal{y}}$& - &  0.0190 & 0.0125 & - & 0.0444 & 0.0265 \\ \hline
CFG-NIN    & 0.0362 & 0.0165 & 0.0071 & 0.0404 & 0.0440 & 0.0273 \\ \hline
CFG-NIN$^{\mathcal{z}}$& 0.0189 & 0.0226 & 0.0163 & 0.0246 & 0.0358 & 0.0240 \\ \hline \hline
\end{tabular}
\caption{KL divergence measurements in rawRGB space by using the EM-1. Note that N2N Flow~\cite{maleky2022noise2noiseflow} was trained on SIDD Full dataset. ${\mathcal{y}}$: numbers are taken from \cite{maleky2022noise2noiseflow}. ${\mathcal{z}}$: we report the results by using the training set in EM-2 to cover all ISO levels and use the evaluation metric in EM-1.}
\label{table:test_raw}
\end{table*}

\textbf{rawRGB noise modeling.} Table \ref{table:test_raw} shows the performance of noise synthesis in rawRGB space. The results indicate that our proposed CFG-NIN achieved comparable performance to the specialized network for raw space, meaning that our method has reliable generalization capability on both sRGB and rawRGB space noise modeling. 

\textbf{Denoising with synthetic noise.} To better evaluate the networks, we use the synthetic data to train one denoiser and compare its performance with the one trained on the real data. To compare with the existing methods, we use DnCNN~\cite{zhang2017beyond} which has been used in  ~\cite{maleky2022noise2noiseflow, abdelhamed2019noise, kousha2022modeling, jang2021c2n} as the denoiser and SIDD Medium~\cite{abdelhamed2018high} as the training set. We use \textbf{SIDD benchmark}~\cite{abdelhamed2018high} as the testing set and provide results on both sRGB and rawRGB denoising tasks. The results are evaluated by the peak signal-to-noise ratio (PSNR). As shown in Table~\ref{table:denoising}, the denoiser trained on CFG-NIN synthetic noise achieved better performance than the one trained with sRGB Flow~\cite{kousha2022modeling}, indicating that the distribution of the CFG-NIN synthetic noise is closer to the real one. For rawRGB DnCNN denoiser, CFG-NIN achieved even better performance than the one trained with the Real noisy dataset. For a more thorough comparison and discussions, we provide additional advanced denoiser (NAF~\cite{chen2022simple} and Restormer~\cite{zamir2022restormer}) results in Appendix~\ref{appendix:denoising}. 

\begin{table}[]
    \small
    \centering
    \begin{tabular}{ccc}
    \hline\hline
        Method     & sRGB & rawRGB \\ \hline
        sRGB Flow\cite{kousha2022modeling}& 34.74$^{\mathcal{x}}$ & - \\ \hline
        % NIN& 35.89 & 46.75 \\
        CFG-NIN & 36.07 & 46.64 \\
        CFG-NIN$^{\mathcal{y}}$& 36.03 & 47.41 \\\hline
        Real& 36.60 & 47.06 \\ \hline\hline
    \end{tabular}
    \caption{Comparison of denoising performance (PSNR) between DnCNN~\cite{zhang2017beyond} trained on real and synthetic noise, the higher the better. ${\mathcal{x}}$: the result is taken from \cite{kousha2022modeling}. ${\mathcal{y}}$: the synthetic noise is generated by the CFG-NIN trained on EM-2.}
    \label{table:denoising}
\end{table}

\subsection{Ablation Studies}
\label{subsec:ablations}
\begin{table*}[]
    \centering
    \small
    \begin{tabular}{c|cccccc|c}
    \hline\hline
    \multirow{2}{*}{Method} & \multicolumn{6}{c|}{$D_{KL}\downarrow$(EM-1, rare ISO levels excluded)} & \multirow{2}{*}{\#Param} \\ \cline{2-7}
     & G4 & GP & IP & N6 & S6 & Agg &  \\ \hline
    sRGB Flow~\cite{kousha2022modeling} & 0.044 & 0.059 & 0.020 & 0.062 & 0.050 & 0.044 & 6.160K \\ \hline
    CFG-NIN-5 & 0.046 & 0.020 & 0.016 & 0.050 & 0.051 & \textbf{0.035} & \textbf{5.256K} \\ \hline
    CFG-NIN-6-LPIPS & 0.040 & 0.029 & 0.022 & 0.058 & 0.065 & 0.043 & \multirow{5}{*}{7.048K} \\
    CFG-NIN-6-GM-Noise & 0.045 & 0.029 & 0.025 & 0.059 & \textbf{0.042} & 0.037 & \\
    CFG-NIN-6-GM-VGG1 & 0.050 & 0.032 & 0.031 & 0.063 & 0.054 & 0.044 & \\
    CFG-NIN-6-GM-VGG12 & 0.043 & 0.028 & 0.018 & 0.050 & 0.043 & 0.034 & \\
    CFG-NIN-6~(-GM-VGG12345) & \textbf{0.032} &  \textbf{0.020} & \textbf{0.018} & \textbf{0.046} & 0.044 & \textbf{0.031} & \\ \hline
    CFG-NIN-6-w/o-CM & 0.082 & 0.125 & 0.063 & 0.158 & 0.060 & 0.086 & 6.940K \\ \hline\hline
    \end{tabular}
    \caption{Ablation studies on the number of parameters, $\mathcal{L}_{style}$, and control map (CM). Note that sRGB Flow~\cite{kousha2022modeling} hard-coded the control information inside the invertible network. Namely, both sRGB Flow and our CFG-NIN-5 utilize control mechanism to achieve better quality.}
    \label{table:ablation_2}
\end{table*}

We benchmark several variations of our CFG-NIN in order to motivate our design choices and to better understand their impact. We focus on noise modeling in sRGB space for all the quantitative comparison in this section. See Appendix~\ref{appendix:ablations_arch} for additional ablation studies on the network architecture.

\textbf{Number of Parameters.} Our prime model, CFG-NIN's number of parameters is lower than all the GAN-based methods' mentioned in Table~\ref{table:test_srgb}. To further validate the effectiveness of our network's design, we cut the number of feature channels from 96 to 6 and 5 to match the parameter count of sRGB Flow, denoted as \textit{CFG-NIN-6} and \textit{CFG-NIN-5}, and trained them with patch size $32\times 32$. As shown in Table~\ref{table:ablation_2}, \textit{CFG-NIN-5} achieved lower $D_{KL}$ than sRGB Flow with fewer parameters.

\textbf{Importance of the Control Information.} Fig.~\ref{fig:var_vs_intensity} illustrates the discrepancies among noise distributions under various camera settings. This highlights the necessity of setting awareness in generative models, which was analyzed as one ablation study in~\cite{kousha2022modeling}. Both sRGB Flow \cite{kousha2022modeling} and our CFG-NIN leverage the availability of control information to effectively model complex distributions,  whereas the curves of C2N remain largely flat across most of the intensity levels. Ensuring completeness, we present a variant of our model trained without the concatenation of the control map, denoted as \emph{CFG-NIN-6-w/o-CM} in Table~\ref{table:ablation_2}. Notably, NeCA~\cite{fu2023srgb} trained individual noise models for each camera and employed a separate model (GENet) to predict the gain factor from a noisy reference image. However, as shown in Table~\ref{table:test_srgb} and Fig.~\ref{fig:comp_visual}, we demonstrated that a unified model trained with control information achieved superior performance and flexibility than the NeCA~\cite{fu2023srgb} approach.

\begin{figure}[]
    \centering
    \subcaptionbox{Comparison between the models trained with/without $\mathcal{L}_{style}$.\label{fig:artifacts_1}}{%
        \includegraphics[width=\linewidth]{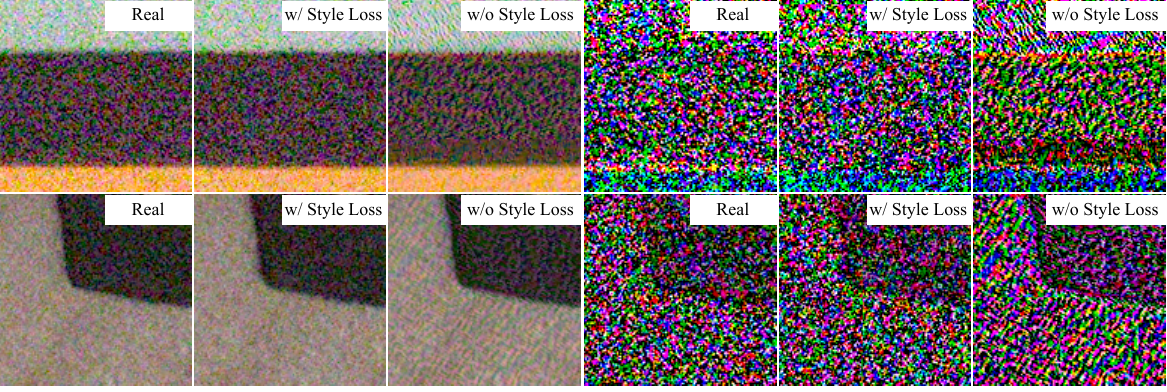}%
    }
    \enskip
    \subcaptionbox{Effect of the number of VGG features being used in $\mathcal{L}_{style}$.\label{fig:artifacts_2}}{%
        \includegraphics[width=\linewidth]{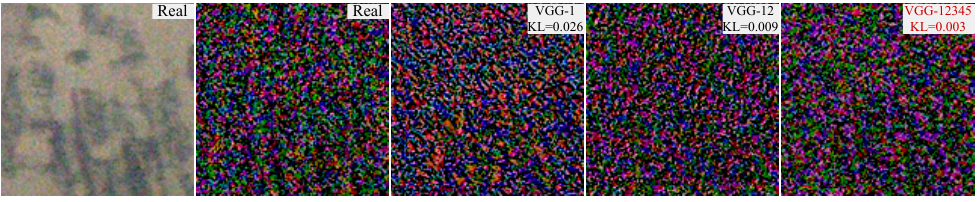}%
    }
    \enskip
    \subcaptionbox{Comparison between $\mathcal{L}_{style}$ applied in sRGB space/perceptual feature space.\label{fig:artifacts_3}}{%
        \includegraphics[width=\linewidth]{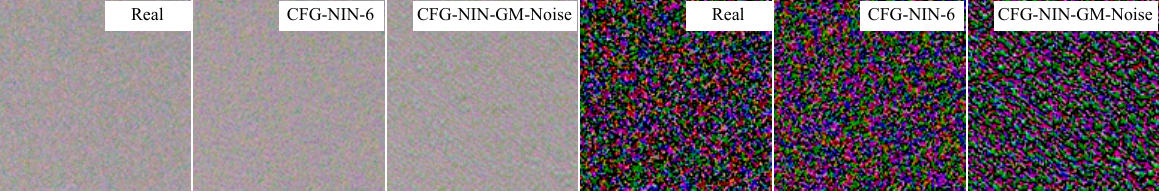}%
    }
    \caption{Visual comparisons of ablation studies on $\mathcal{L}_{style}$.}
    \label{fig:artifacts}
\end{figure}

\textbf{Importance of the Style Loss.} To demonstrate the effectiveness of the $\mathcal{L}_{style}$ in the noise modeling task, we trained CFG-NIN only with the adversarial loss from the discriminator. As shown in Fig.~\ref{fig:artifacts_1}, the synthesized noise (without $\mathcal{L}_{style}$) contains repetitive artifacts. 

To explore other potential non-adversarial supervisions, we substituted $\mathcal{L}_{style}$ with perceptual loss, LPIPS, denoted as \emph{CFG-NIN-6-LPIPS} in Table~\ref{table:ablation_2}. Similar to other pixel-wise losses (e.g., MSE), LPIPS is deterministic and we observed its negative effect on temporal variance. 

To assess the impact of multiscale deep features in LPIPS measurement, we reduce the number of layers $L$ in Eq.~\ref{eq:style_loss}, denoted as \emph{CFG-NIN-6-GM-VGG\textbf{(l)}} in Table~\ref{table:ablation_2}. Both quantitative and visual comparisons shown in Fig.~\ref{fig:artifacts_2} demonstrate the crucial role of MMD over multiscale features in improving performance. We interpret that the incorporation of multi-resolution features in VGG layers facilitates the capture of finer local correlations. This is corroborated by the findings of the \textit{CFG-NIN-6-GM-Noise} variant, as shown in Fig.~\ref{fig:artifacts_3}. In this variant, we apply the GM on the generated noise map in the sRGB space, and its results contain similar granularity artifacts to those of \textit{CFG-NIN-6-GM-VGG1} and the model trained without $\mathcal{L}_{style}$.

\subsection{Qualitative Results} 
\label{subsec:visual}
\begin{figure}
    \centering
    \includegraphics[width=0.95\linewidth]{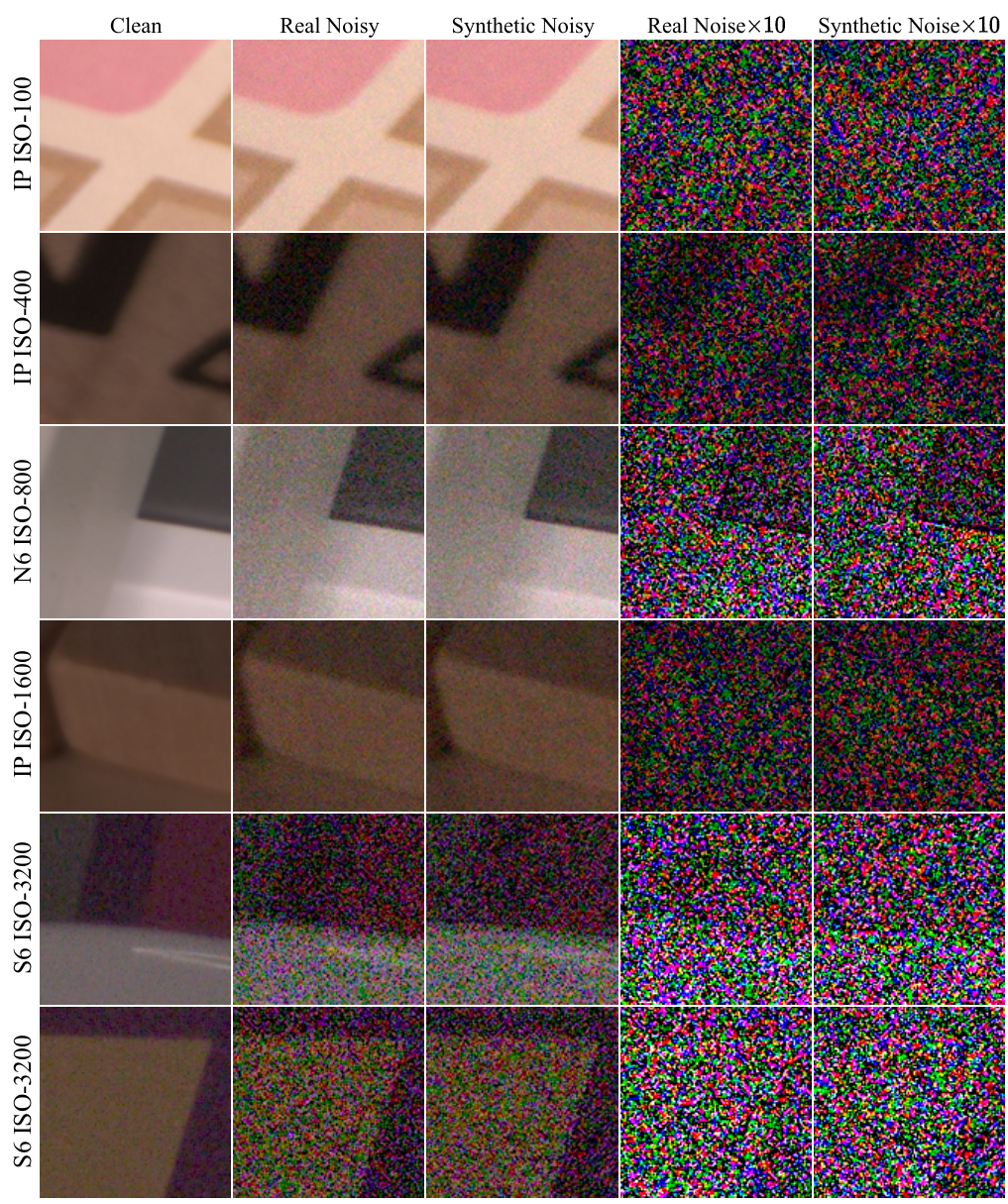}
    \caption{The synthetic noisy images and noise maps on the sRGB noise modeling task.}
    \label{fig:visual_rgb}
\end{figure}
\begin{figure}
    \centering
    \includegraphics[width=0.95\linewidth]{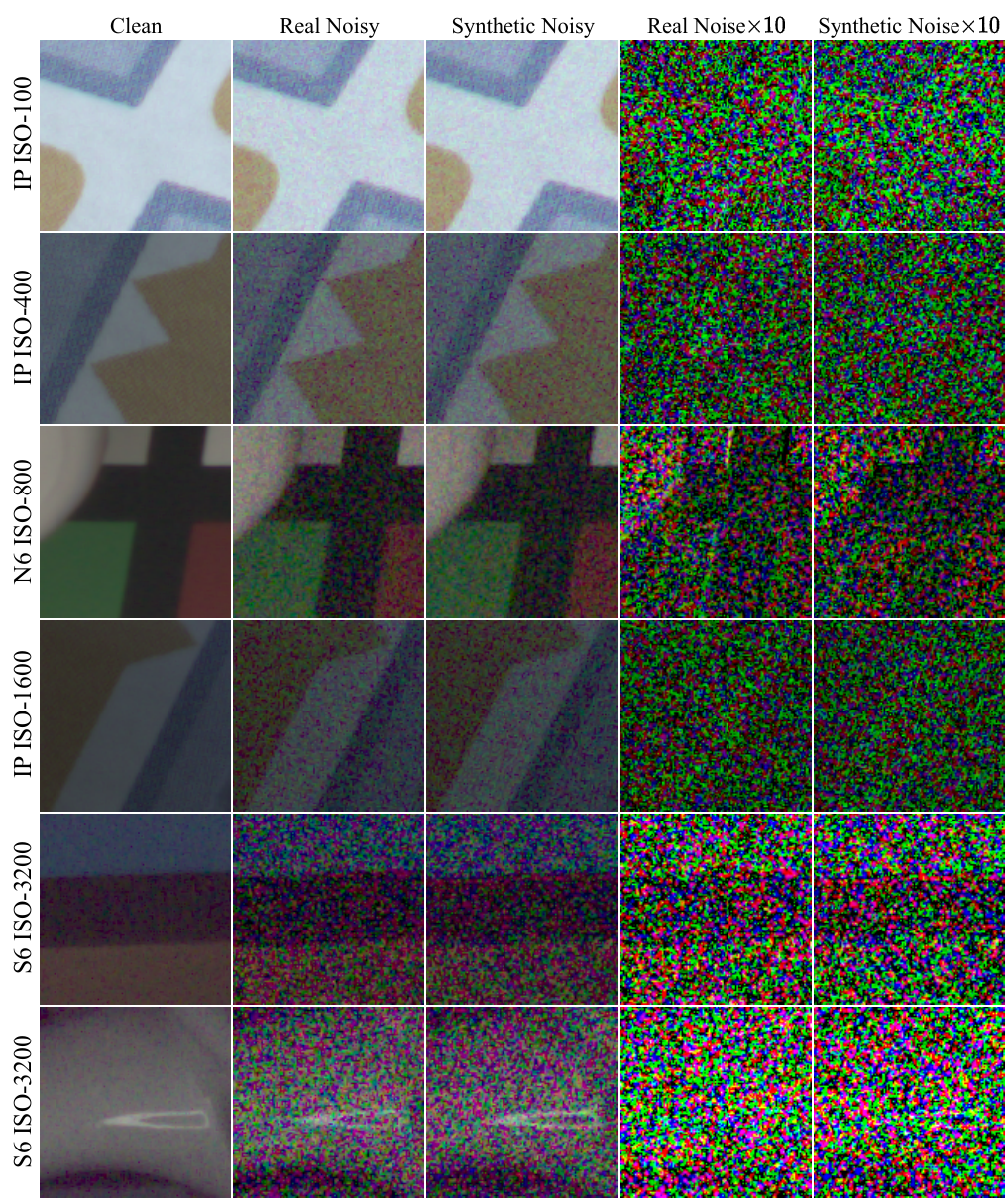}
    \caption{The synthetic noisy images and noise maps on the rawRGB noise modeling task. The images are converted into sRGB space for better visualization.}
    \label{fig:visual_raw}
\end{figure}

Fig.~\ref{fig:visual_rgb} and \ref{fig:visual_raw} show the qualitative results of the synthesized noise by our proposed model (trained on the list\textsuperscript{\ref{note1}} provided by ~\cite{maleky2022noise2noiseflow}) on sRGB and rawRGB noise modeling tasks. To visualize the rawRGB images better, we use the scripts provided in ~\cite{abdelhamed2019noise} to convert them into sRGB color space and multiply the noise map by 10 for better visualization. We provide the results produced by the model trained on EM-2 in Appendix~\ref{appendix:visual}, which contains higher ISO levels.

\subsection{Spatial Correlation} 
\label{subsec:correlate}
\begin{figure}
    \centering
    \subcaptionbox{\label{fig:correlation_position}}{%
        \includegraphics[height=3.4cm]{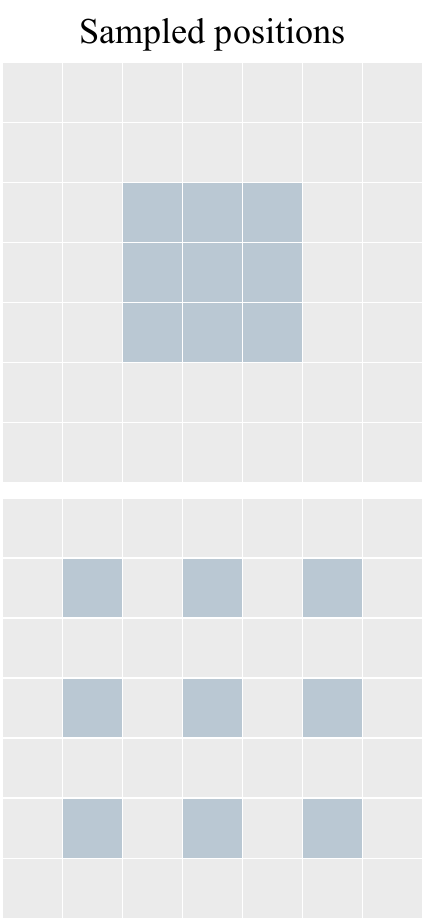}%
    }
    % \enskip
    \subcaptionbox{sRGB\label{fig:correlation_srgb}}{%
        \includegraphics[height=3.4cm]{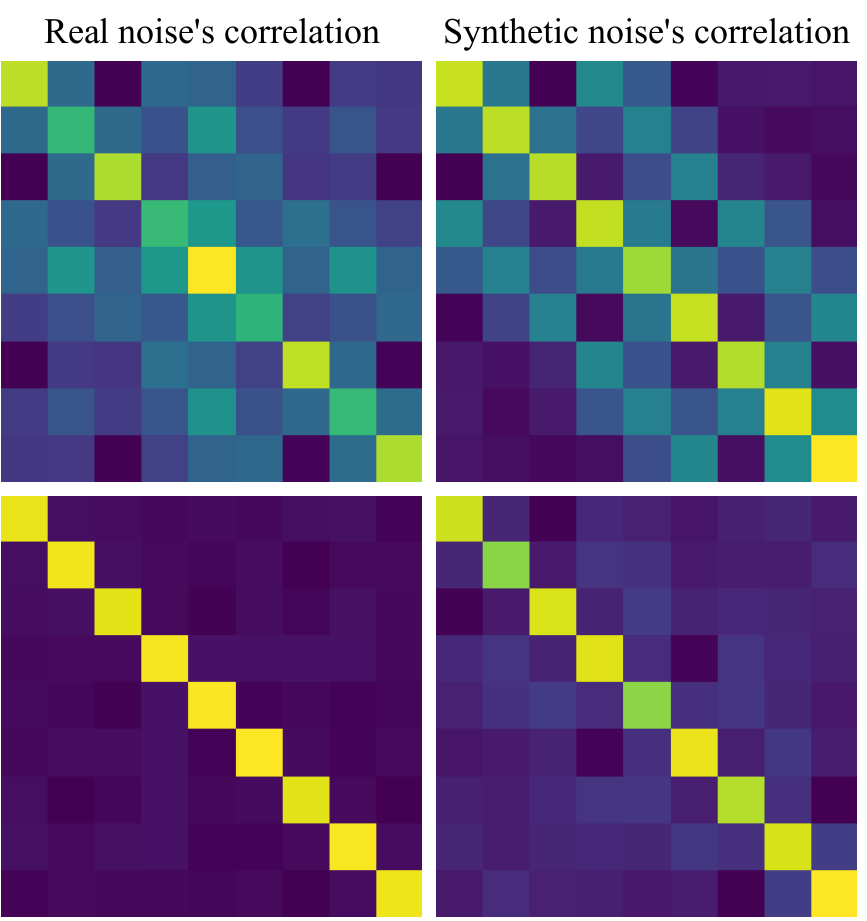}%
    }
    % \enskip
    \subcaptionbox{rawRGB\label{fig:correlation_raw}}{%
        \includegraphics[height=3.4cm]{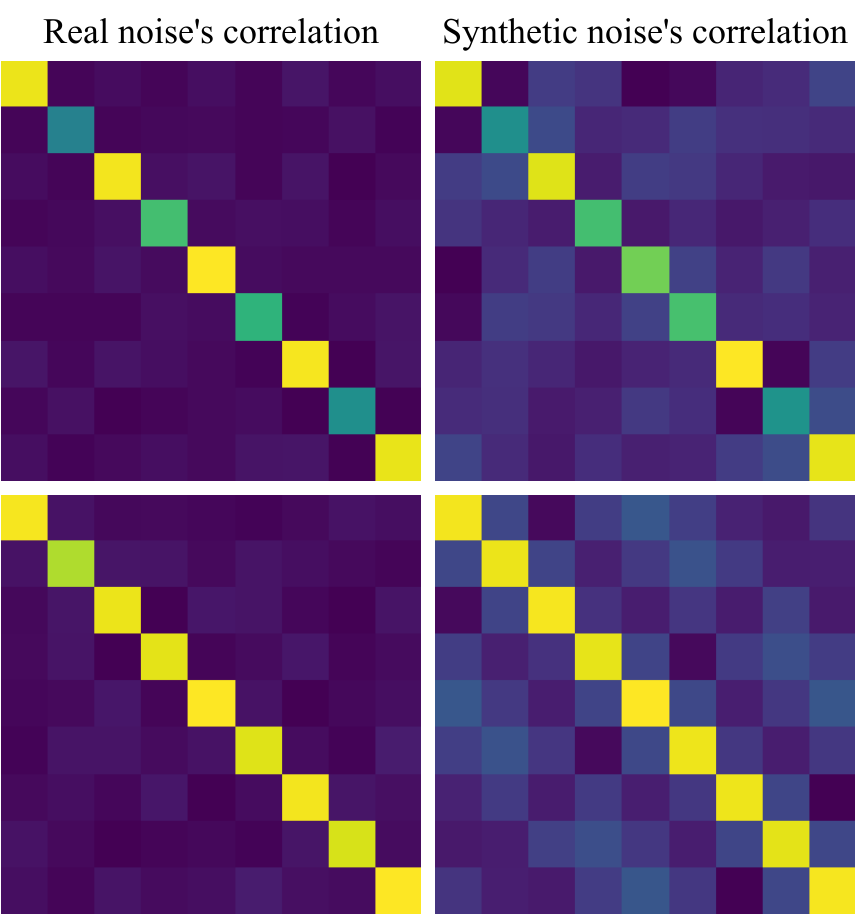}%
    }
    \caption{Spatial correlation of the noises sampled on pixels with different distances from each other (shown in (a)) in sRGB space (b) and rawRGB space (c). 
    % For comparison in (b) and (c), the left column shows real noise's correlation and the right column shows the synthetic noise's. Brighter parts represent higher covariance.
    }
    \label{fig:correlation}
\end{figure}

We visualize the correlation of the noises w.r.t relative spatial position by enlarging the distance between sampled pixels from one to two and collecting the noise values on these pixels. As shown in Fig~\ref{fig:correlation_srgb}, the covariance between different pixels approaches to zero when the distance is larger than one, and our proposed network can produce a similar relationship between adjacent pixels.

As shown in Fig~\ref{fig:correlation_raw}, the noises in rawRGB space show little correlation even on neighboring pixels with a distance of one. With the same proposed architecture as the network trained in sRGB space, the network trained on rawRGB domain can generalize to such different behavior. Comparing the results shown in Fig~\ref{fig:correlation_srgb} and ~\ref{fig:correlation_raw}, we interpreted that the correlation in sRGB space is introduced by the demosaicing in ISP, which indicates that independently sampling noise in rawRGB space is sufficient to model the realistic noise distribution, but inadequate for sRGB space noise synthesis.

\subsection{Temporal Variance} 
\label{subsec:temporal_variance}
\begin{figure}[]
    \centering
    \includegraphics[width=0.95\linewidth]{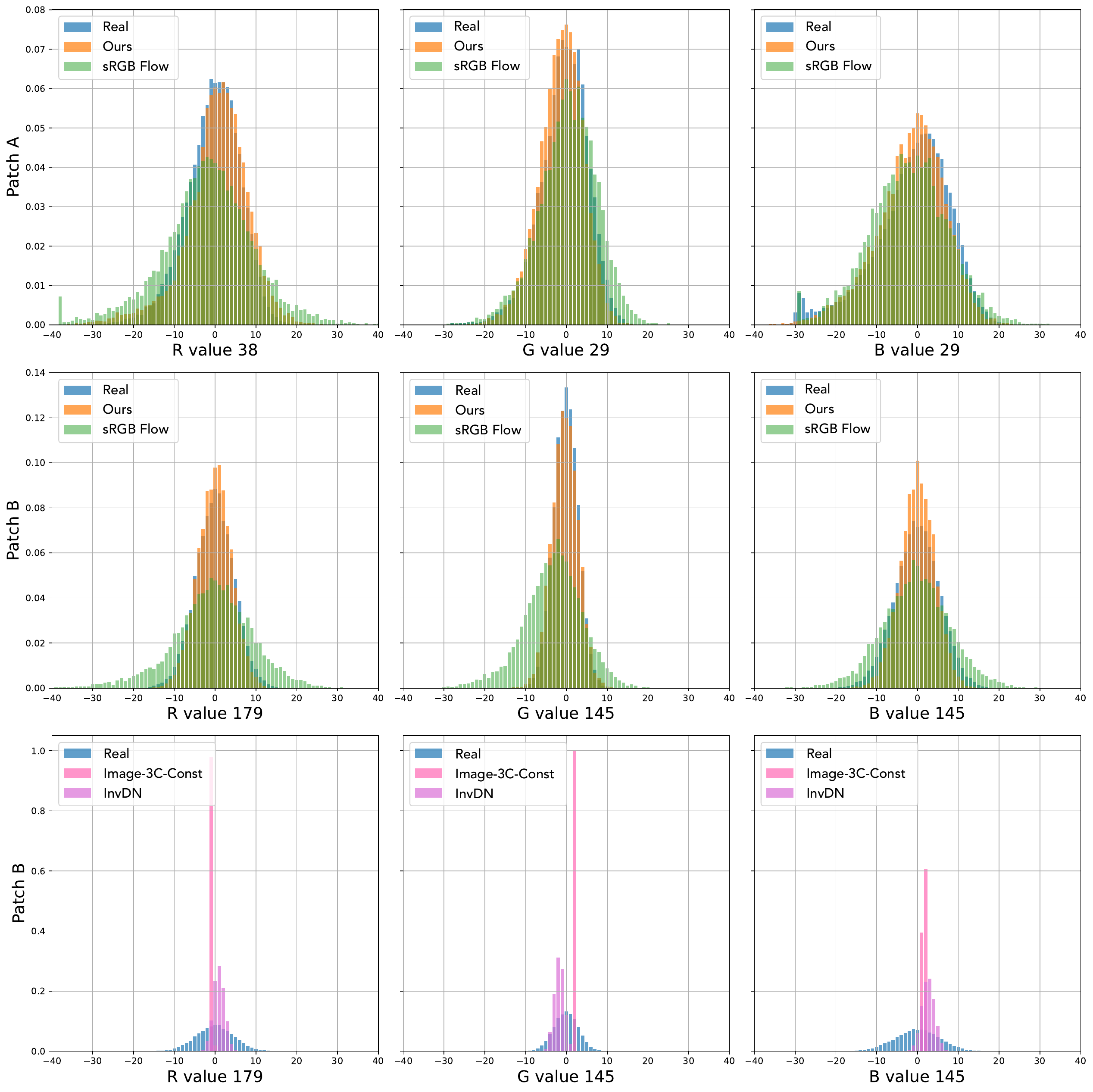}
    \caption{The histogram of the sampled noises on the pixel with a specific value in sRGB space. The columns demonstrate R, G and B channels respectively.}
    \label{fig:temporal_hist}
\end{figure}

The issues of the empirical statistical results produced by sampling the testing set only once, which cannot capture the whole picture of the model that learns the real noise distribution. Therefore, to simulate the actual process of generating noise, we choose one flat region on an image and sample the noise on this patch 10000 times. Then we choose the pixel in the center of the patch and calculate the empirical frequency distribution histogram of the noise on it. For the ground truth of such histogram, we use SIDD-Full~\cite{abdelhamed2018high} dataset, where each clean image has 150 pairing noisy ones (see Appendix~\ref{appendix:sample_remporal} for detailed sampling strategy). To construct a benchmark for comparison, we remove the spatial variation of noise and only sample $C=3$ random values, one for each channel, and then replicate them to fill the shape $W \times H$. In this case, the noise value is identical in each spatial location of a channel and only varies across channels. This model is referred to as \emph{Image-3C-Const}. Fig.~\ref{fig:temporal_hist} shows the comparison of different methods on the sRGB noise modeling task. As shown in the histograms, our network correctly synthesizes the randomness of the noise with the same variances as the real one.

The third row of Fig.~\ref{fig:temporal_hist} indicates that, although \textit{Image-3C-Const} generated samples have close distribution to the real one (see Table~\ref{table:appendix_rgb_results}-\ref{table:appendix_rgb_results_EM2} in Sec.~\ref{appendix:quantitative} for more details), it is false to imitate the actual temporal randomness of the noise. More specifically, although the KL divergence between the real noise map and the synthesized one by \textit{Image-3C-Const} is small, however, given different random seeds, the static appearance of the noise that produces strong dragging artifacts on the re-noised video frames by \textit{Image-3C-Const}. Namely, the noises move with the actual image content without any temporal randomness. As a result, we call the randomness of synthetic noise w.r.t the random seed the temporal variance.

\begin{figure}[]
    \centering
    \includegraphics[width=0.95\linewidth]{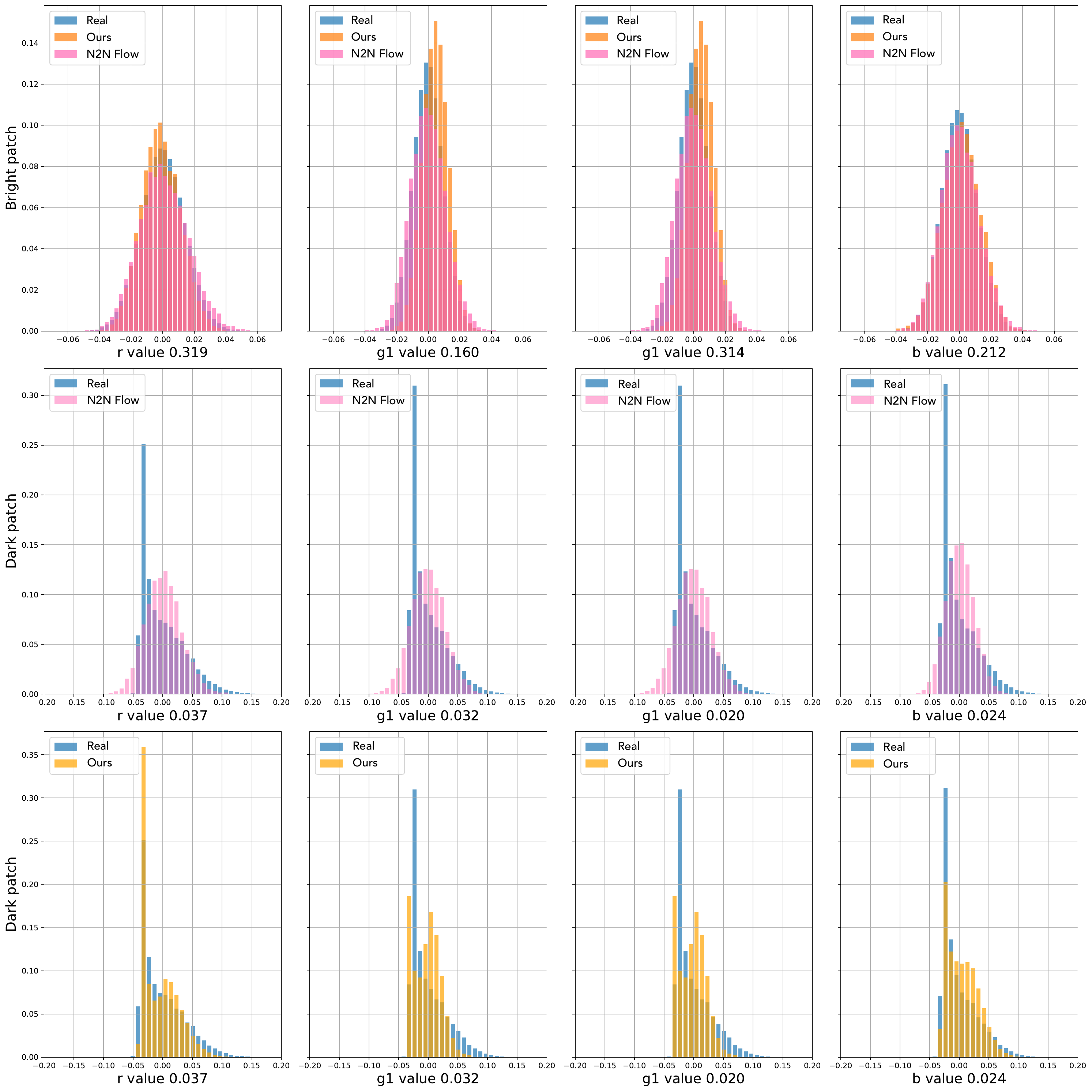}
    \caption{The histogram of the sampled noises on the pixel with a specific value in rawRGB space. The value of each clean channel is shown as R, G1, G2, B, respectively, in the figure. The first row demonstrates the performance of the bright patch, and the rest two rows for the dark patch.}
    \label{fig:temporal_hist_raw}
\end{figure}

Fig.~\ref{fig:temporal_hist_raw} shows the temporal variance of the noise generated by our model on rawRGB domain. As shown in the middle and bottom rows, the real noise distribution is similar to the Poisson distribution on pixels with low intensity, which is consistent with the hypothesis of shot noise, and our model can generate noises closer to such distribution (shown in Fig.~\ref{fig:temporal_hist_raw} bottom row).

\section{Conclusion}
\label{sec:conslusion}
In this work, we proposed an effective model for synthesizing realistic and controllable digital camera noise that outperforms existing methods in both sRGB and rawRGB space. Our approach is based on a conditional GAN training scheme, where a noise generator is constructed by concatenating the Gaussian noise map to the transition point between the encoder and decoder and injecting Gaussian noise into each decoder block. Our proposed method keeps the encoded features clean and utilizes them as guidance for modeling the content-dependent characteristics of the noise. We employ a Style Loss to supervise the generator training, which resolves the instability issues of GAN training while maintaining the crucial stochastic properties of the synthetic noise. We quantitatively demonstrated that our generated noise exhibits more accurate temporal variance and spatial correlation than existing methods, indicating the effectiveness of our approach for synthesizing digital camera noise.

{\small
\bibliographystyle{ieee_fullname}
\bibliography{egbib}

\begin{thebibliography}{10}\itemsep=-1pt

\bibitem{abdelhamed2019noise}
Abdelrahman Abdelhamed, Marcus~A Brubaker, and Michael~S Brown.
\newblock Noise flow: Noise modeling with conditional normalizing flows.
\newblock In {\em Proceedings of the IEEE/CVF International Conference on Computer Vision}, pages 3165--3173, 2019.

\bibitem{abdelhamed2018high}
Abdelrahman Abdelhamed, Stephen Lin, and Michael~S Brown.
\newblock A high-quality denoising dataset for smartphone cameras.
\newblock In {\em Proceedings of the IEEE Conference on Computer Vision and Pattern Recognition}, pages 1692--1700, 2018.

\bibitem{ameur2022deep}
Zoubida Ameur, Wassim Hamidouche, Edouard Fran{\c{c}}ois, Milo{\v{s}} Radosavljevi{\'c}, Daniel Menard, and Claire-H{\'e}l{\`e}ne Demarty.
\newblock Deep-based film grain removal and synthesis.
\newblock {\em arXiv preprint arXiv:2206.07411}, 2022.

\bibitem{ba2016layer}
Jimmy~Lei Ba, Jamie~Ryan Kiros, and Geoffrey~E Hinton.
\newblock Layer normalization.
\newblock {\em arXiv preprint arXiv:1607.06450}, 2016.

\bibitem{cai2021learning}
Yuanhao Cai, Xiaowan Hu, Haoqian Wang, Yulun Zhang, Hanspeter Pfister, and Donglai Wei.
\newblock Learning to generate realistic noisy images via pixel-level noise-aware adversarial training.
\newblock {\em Advances in Neural Information Processing Systems}, 34:3259--3270, 2021.

\bibitem{chen2022simple}
Liangyu Chen, Xiaojie Chu, Xiangyu Zhang, and Jian Sun.
\newblock Simple baselines for image restoration.
\newblock {\em arXiv preprint arXiv:2204.04676}, 2022.

\bibitem{fu2023srgb}
Zixuan Fu, Lanqing Guo, and Bihan Wen.
\newblock srgb real noise synthesizing with neighboring correlation-aware noise model.
\newblock In {\em Proceedings of the IEEE/CVF Conference on Computer Vision and Pattern Recognition}, pages 1683--1691, 2023.

\bibitem{gatys2016image}
Leon~A Gatys, Alexander~S Ecker, and Matthias Bethge.
\newblock Image style transfer using convolutional neural networks.
\newblock In {\em Proceedings of the IEEE conference on computer vision and pattern recognition}, pages 2414--2423, 2016.

\bibitem{goodfellow2020generative}
Ian Goodfellow, Jean Pouget-Abadie, Mehdi Mirza, Bing Xu, David Warde-Farley, Sherjil Ozair, Aaron Courville, and Yoshua Bengio.
\newblock Generative adversarial networks.
\newblock {\em Communications of the ACM}, 63(11):139--144, 2020.

\bibitem{hu2018squeeze}
Jie Hu, Li Shen, and Gang Sun.
\newblock Squeeze-and-excitation networks.
\newblock In {\em Proceedings of the IEEE conference on computer vision and pattern recognition}, pages 7132--7141, 2018.

\bibitem{jang2021c2n}
Geonwoon Jang, Wooseok Lee, Sanghyun Son, and Kyoung~Mu Lee.
\newblock {C2N}: Practical generative noise modeling for real-world denoising.
\newblock In {\em Proceedings of the IEEE/CVF International Conference on Computer Vision}, pages 2350--2359, 2021.

\bibitem{karras2020analyzing}
Tero Karras, Samuli Laine, Miika Aittala, Janne Hellsten, Jaakko Lehtinen, and Timo Aila.
\newblock Analyzing and improving the image quality of stylegan.
\newblock In {\em Proceedings of the IEEE/CVF conference on computer vision and pattern recognition}, pages 8110--8119, 2020.

\bibitem{kousha2022modeling}
Shayan Kousha, Ali Maleky, Michael~S Brown, and Marcus~A Brubaker.
\newblock Modeling {sRGB} camera noise with normalizing flows.
\newblock In {\em Proceedings of the IEEE/CVF Conference on Computer Vision and Pattern Recognition}, pages 17463--17471, 2022.

\bibitem{lehtinen2018noise2noise}
Jaakko Lehtinen, Jacob Munkberg, Jon Hasselgren, Samuli Laine, Tero Karras, Miika Aittala, and Timo Aila.
\newblock Noise2noise: Learning image restoration without clean data.
\newblock {\em arXiv preprint arXiv:1803.04189}, 2018.

\bibitem{li2017demystifying}
Yanghao Li, Naiyan Wang, Jiaying Liu, and Xiaodi Hou.
\newblock Demystifying neural style transfer.
\newblock {\em arXiv preprint arXiv:1701.01036}, 2017.

\bibitem{liang2021swinir}
Jingyun Liang, Jiezhang Cao, Guolei Sun, Kai Zhang, Luc Van~Gool, and Radu Timofte.
\newblock {SwinIR}: Image restoration using swin transformer.
\newblock In {\em Proceedings of the IEEE/CVF International Conference on Computer Vision}, pages 1833--1844, 2021.

\bibitem{liu2021invertible}
Yang Liu, Zhenyue Qin, Saeed Anwar, Pan Ji, Dongwoo Kim, Sabrina Caldwell, and Tom Gedeon.
\newblock Invertible denoising network: A light solution for real noise removal.
\newblock In {\em Proceedings of the IEEE/CVF conference on computer vision and pattern recognition}, pages 13365--13374, 2021.

\bibitem{loshchilov2016sgdr}
Ilya Loshchilov and Frank Hutter.
\newblock {SGDR}: Stochastic gradient descent with warm restarts.
\newblock {\em arXiv preprint arXiv:1608.03983}, 2016.

\bibitem{maleky2022noise2noiseflow}
Ali Maleky, Shayan Kousha, Michael~S Brown, and Marcus~A Brubaker.
\newblock Noise2noiseflow: Realistic camera noise modeling without clean images.
\newblock In {\em Proceedings of the IEEE/CVF Conference on Computer Vision and Pattern Recognition}, pages 17632--17641, 2022.

\bibitem{monakhova2022dancing}
Kristina Monakhova, Stephan~R Richter, Laura Waller, and Vladlen Koltun.
\newblock Dancing under the stars: video denoising in starlight.
\newblock In {\em Proceedings of the IEEE/CVF Conference on Computer Vision and Pattern Recognition}, pages 16241--16251, 2022.

\bibitem{norkin2018film}
Andrey Norkin and Neil Birkbeck.
\newblock Film grain synthesis for {AV1} video codec.
\newblock In {\em 2018 Data Compression Conference}, pages 3--12. IEEE, 2018.

\bibitem{rezende2015variational}
Danilo Rezende and Shakir Mohamed.
\newblock Variational inference with normalizing flows.
\newblock In {\em International conference on machine learning}, pages 1530--1538. PMLR, 2015.

\bibitem{shazeer2020glu}
Noam Shazeer.
\newblock {GLU} variants improve transformer.
\newblock {\em arXiv preprint arXiv:2002.05202}, 2020.

\bibitem{simonyan2014very}
Karen Simonyan and Andrew Zisserman.
\newblock Very deep convolutional networks for large-scale image recognition.
\newblock {\em arXiv preprint arXiv:1409.1556}, 2014.

\bibitem{tassano2020fastdvdnet}
Matias Tassano, Julie Delon, and Thomas Veit.
\newblock Fastdvdnet: Towards real-time deep video denoising without flow estimation.
\newblock In {\em Proceedings of the IEEE/CVF Conference on Computer Vision and Pattern Recognition}, pages 1354--1363, 2020.

\bibitem{wei2020physics}
Kaixuan Wei, Ying Fu, Jiaolong Yang, and Hua Huang.
\newblock A physics-based noise formation model for extreme low-light raw denoising.
\newblock In {\em Proceedings of the IEEE/CVF Conference on Computer Vision and Pattern Recognition}, pages 2758--2767, 2020.

\bibitem{yue2020dual}
Zongsheng Yue, Qian Zhao, Lei Zhang, and Deyu Meng.
\newblock Dual adversarial network: Toward real-world noise removal and noise generation.
\newblock In {\em Computer Vision--ECCV 2020: 16th European Conference, Glasgow, UK, August 23--28, 2020, Proceedings, Part X 16}, pages 41--58. Springer, 2020.

\bibitem{zamir2022restormer}
Syed~Waqas Zamir, Aditya Arora, Salman Khan, Munawar Hayat, Fahad~Shahbaz Khan, and Ming-Hsuan Yang.
\newblock Restormer: Efficient transformer for high-resolution image restoration.
\newblock In {\em Proceedings of the IEEE/CVF Conference on Computer Vision and Pattern Recognition}, pages 5728--5739, 2022.

\bibitem{zhang2017beyond}
Kai Zhang, Wangmeng Zuo, Yunjin Chen, Deyu Meng, and Lei Zhang.
\newblock Beyond a gaussian denoiser: Residual learning of deep cnn for image denoising.
\newblock {\em IEEE transactions on image processing}, 26(7):3142--3155, 2017.

\bibitem{zhang2021rethinking}
Yi Zhang, Hongwei Qin, Xiaogang Wang, and Hongsheng Li.
\newblock Rethinking noise synthesis and modeling in raw denoising.
\newblock In {\em Proceedings of the IEEE/CVF International Conference on Computer Vision}, pages 4593--4601, 2021.

\bibitem{zou2022estimating}
Yunhao Zou and Ying Fu.
\newblock Estimating fine-grained noise model via contrastive learning.
\newblock In {\em Proceedings of the IEEE/CVF Conference on Computer Vision and Pattern Recognition}, pages 12682--12691, 2022.

\end{thebibliography}
}

\clearpage
% \appendix
\begin{appendices}

% \twocolumn[{%
% \renewcommand\twocolumn[1][]{#1}%
% \begin{center}
%     \centering
%     \captionsetup{type=figure}
%     \includegraphics[width=0.8\linewidth]{figures/hyper_NN.pdf}
%     \captionof{figure}{Hyperparameters of CFG-NIN. $B$ is the batch size, $C$ is the summation of the clean image channel length and control map channel length, $C^*$ is the channel length of the synthetic noise map, and $H,W$ represent the height and width of the patch.}
%     \label{fig:hyper_nn}
% \end{center}%
% }]

\begin{figure*}
    \centering
    \includegraphics[width=0.8\linewidth]{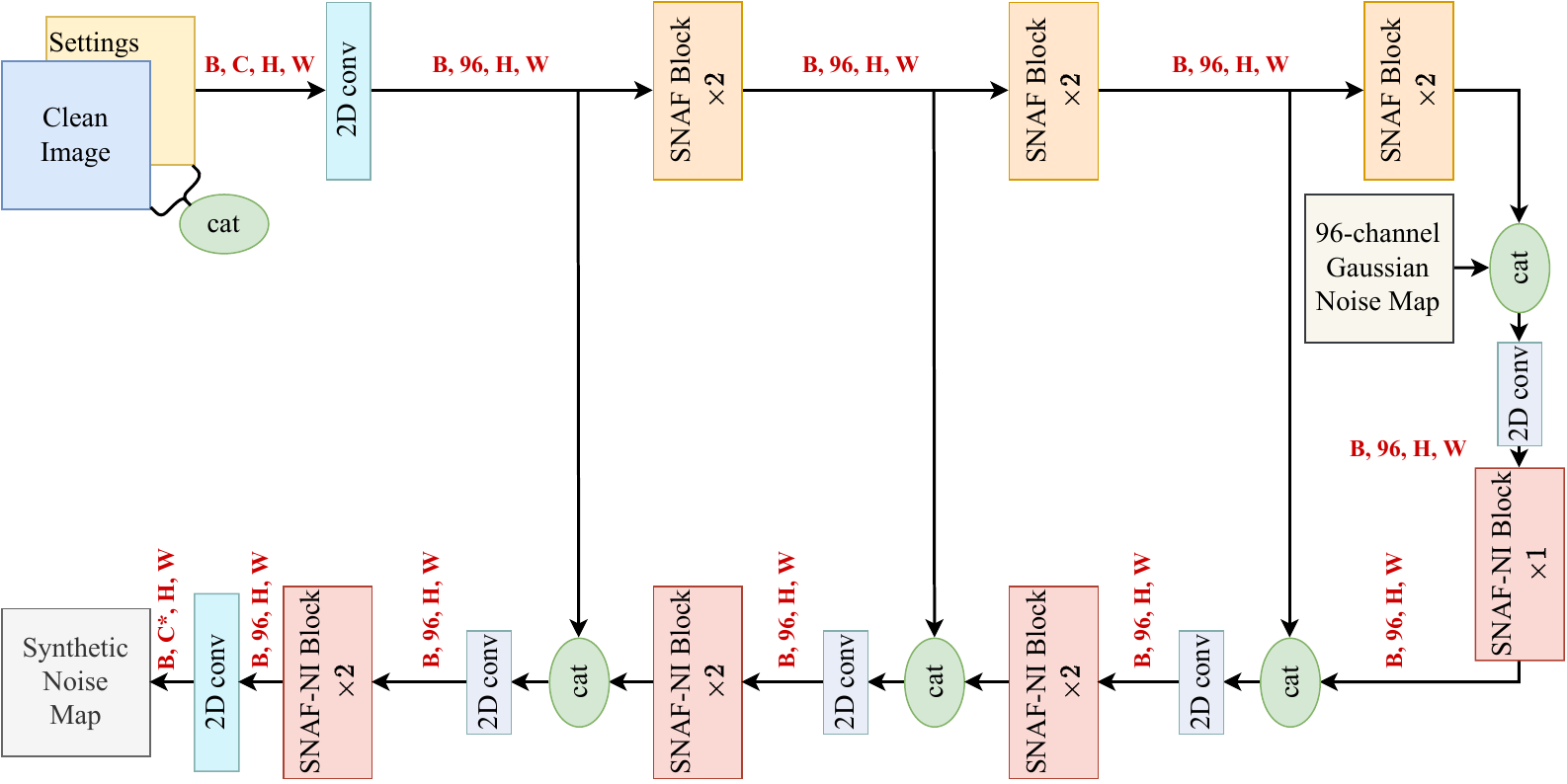}
    \caption{Hyperparameters of CFG-NIN. $B$ is the batch size, $C$ is the summation of the clean image channel length and control map channel length, $C^*$ is the channel length of the synthetic noise map, and $H,W$ represent the height and width of the patch.}
    \label{fig:hyper_nn}
\end{figure*}

\section{Hyperparameters}
\label{appendix:hyper}
\subsection{Network}
\label{appendix:network}
To ensure our work can be easily implemented and the results can be reproduced, we provide the detail of the hyperparameters of our prime model, CFG-NIN. The encoder (clean feature extraction layers) and decoder (noise synthesis layers) described in the main paper (Fig. 2) are sequences of SNAF (SNAF-NI) blocks, where the number of blocks is two (i.e., $n=2$). The channel length of the feature is set to 96 throughout the whole network. To keep consistency, the channel length of the concatenated Gaussian noise at the transition point is set to 96 (i.e., $m=96$). A 2D convolutional layer is applied directly after all the concatenations to reduce the channel length back to 96. We provide the dimension (in \red \textbf{red} \black characters) of the tensor between each layer shown in Fig~\ref{fig:hyper_nn}. The total number of parameters of the CFG-NIN is $1.186\times10^6$. We use the reflection padding for all the 2D convolutional layers in the network.

\subsection{Control Map}
Blind noise synthesizing is intractable due to the ambiguity of the relationship between noise distributions and clean pixel intensities. Moreover, generating noise/grain with different styles on video clips is preferred for artistic control. The discrepancies between noise distributions under different ISO levels and camera brands have been thoroughly analyzed in ~\cite{kousha2022modeling}. For camera settings, e.g. ISO and shutter speed, we normalize them to $[0,1]$. For the camera brand marks, we encode them from string into unique floating numbers (e.g., \lstinline{crc32(string)}). We then broadcast these scalars to 2D and concatenate them to the other inputs along the channel dimension. Such a process can be written in Pytorch style as follows:

\begin{listing}[h]%
\caption{Control Map}%
\label{lst:listing}%
\begin{lstlisting}[language=Python]
cm=[torch.full(
        size=(1,H,W),
        fill_value=c
    ) for c in controls_encoded]
cm=torch.cat(cm,dim=0)
inp=torch.cat([inp,cm],dim=0)
# shape of inp: C, H, W
\end{lstlisting}
\end{listing}
where the \lstinline{controls_encoded} is the aforementioned encoded camera settings and brandmark. Such a conditioning method is widely used in low-level vision tasks and is flexible and compatible compared with hard coding in the network.

\subsection{Losses}
\label{appendix:losses}
As described in Eq.~\ref{eq:disc_loss}-~\ref{eq:adv_loss} in the main paper, we use the Squared Error as discriminator scores, which means that $\mathcal{L}_{disc}$ and $\mathcal{L}_{adv}$ vibrate around $0.5^2$. To balance the adversarial loss and Style Loss and keep them in the same order of magnitude during the early epochs of training, we set the total loss for the generator CFG-NIN as:
\begin{equation}
    \mathcal{L}_{total}=\mathcal{L}_{adv}+\lambda\mathcal{L}_{style},
    \label{eq:total}
\end{equation}
where $\lambda$ is the hyperparameter for balancing and is set to $0.5$ and $1.0$ for noise modeling in sRGB and rawRGB space, respectively.

Following~\cite{abdelhamed2019noise, maleky2022noise2noiseflow}, the images are packed from one-channel to four-channel ($R, G_1, G_2, B$) on noise modeling task in rawRGB space. As a result, the pre-trained VGG-Network~\cite{simonyan2014very}, which is specialized for images with RGB 3 channels, cannot be applied directly for Style loss estimation. To solve this problem, we split the packed images along the channel dimension into two parts so that each part contains three channels, and compute the Style Loss on each of them, which can be described as:
\begin{equation}
    \mathcal{L}_{style}=\left(\mathcal{L}_{style}^{RG_1B}+\mathcal{L}_{style}^{RG_2B}\right)/2.
    \label{eq:raw_style_loss}
\end{equation}

\subsection{Training Prodecure}
\label{appendix:training}

The batch size is set to 8, and the total training step is set to $6\times 10^5$ across all the CFG-NIN and its variations shown in Fig. \ref{fig:ablation_1} and \ref{fig:ablation_2}. For each mini-batch, we crop a smaller $128\times128$ patch from each sampled patch as the input. The learning rate decades from $5\times 10^{-5}$ to $1\times 10^{-7}$ with the Cosine Annealing~\cite{loshchilov2016sgdr} method.

\subsection{Ablation Studies (Networks detail)}
\label{appendix:ablations_arch}
As described in Sec.~\ref{subsec:ablations} in the main paper, we provide several additional variations of our CFG-NIN architecture to reason the design of our generator.

\textbf{Location of Noise Injection.} 
Firstly, we want to analyze the impact of injecting noise in different locations of the network. For this purpose, we start by using regular SNAF blocks without noise injection throughout the whole network. We then compare the outcome when injecting noise in different locations.

\begin{figure}
    \centering
    \begin{subfigure}{\linewidth}
        \includegraphics[width=\linewidth]{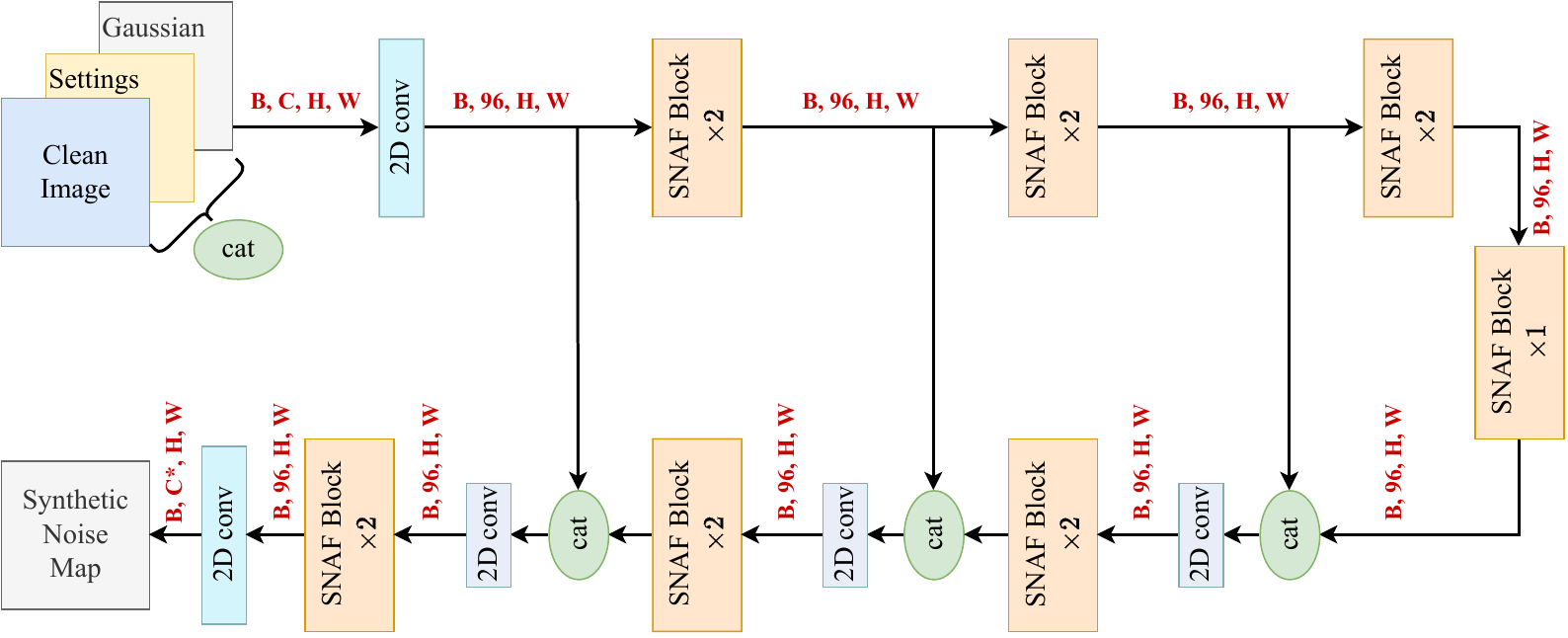}
        \caption{Image-3C, Image-3C-Const and Image-96C}
        \label{fig:cat_rgb}
    \end{subfigure}
    \hfill
    \begin{subfigure}{\linewidth}
        \includegraphics[width=\linewidth]{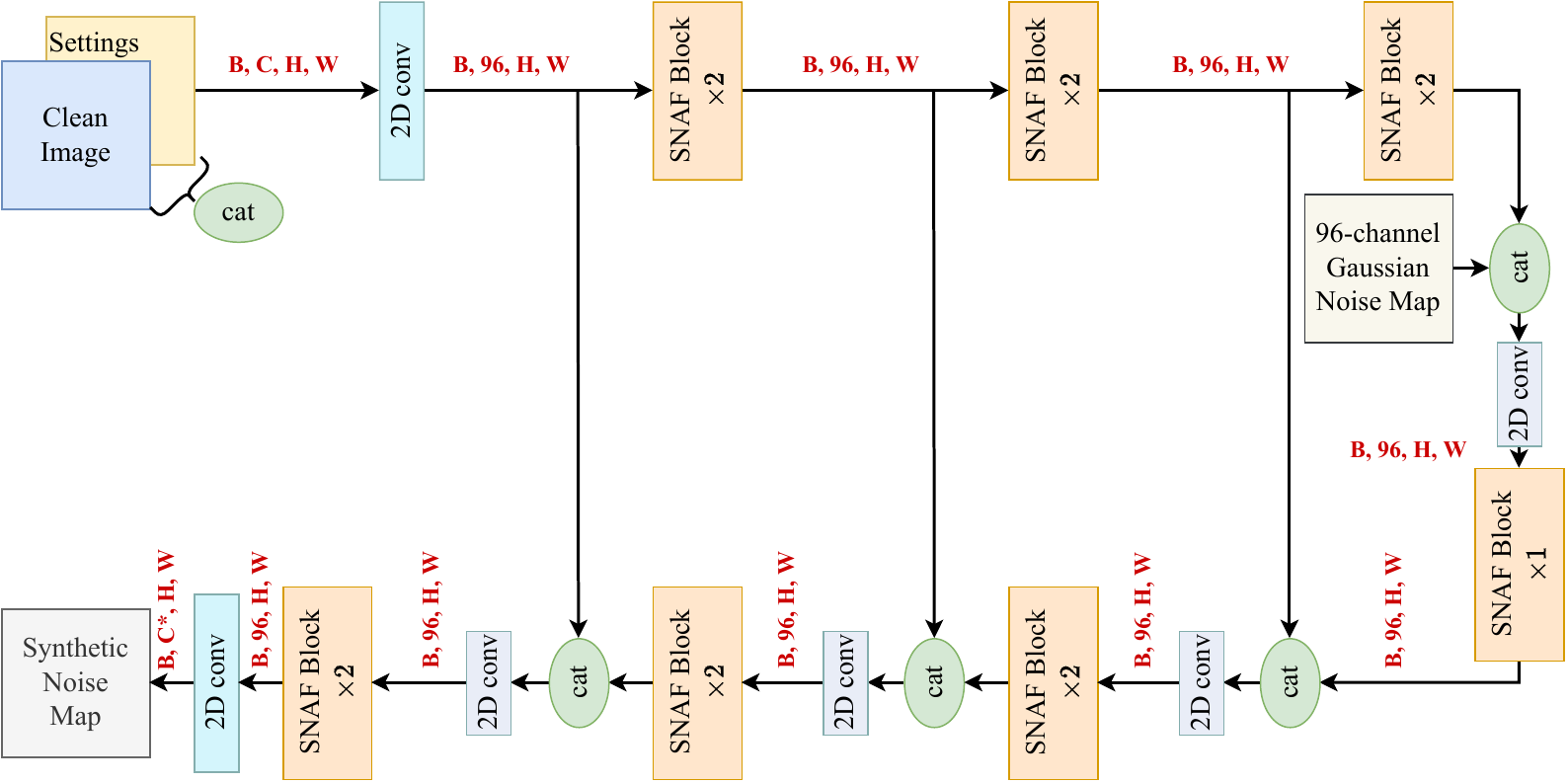}
        \caption{Latent-96C}
        \label{fig:cat_latent}
    \end{subfigure}
    \caption{Ablation study on the position of the concatenated Gaussian noise map.}
    \label{fig:ablation_1}
\end{figure}

Secondly, we inject noise in the latents in the middle of the network. Specifically, we concatenate independently sampled Gaussian noise according to the shape of the latents $(W,H,C=96)$ as shown in Fig.~\ref{fig:cat_latent}. This variation is referred to as \emph{Latent-96C}.

Thirdly, we compare injecting Gaussian noise at the very beginning of the network as in Fig.~\ref{fig:cat_rgb}. This variation is referred to as \emph{Image-3C} since we independently sample according to the size of the input image $(W,H,C=3)$.
In order to compare the effect of the dimensionality of the sampled noise, we also benchmark a variant where we sample 96 channels independently, referred to as \emph{Image-96C}.

Finally, we consider the importance of spatial variation in the injected noise and benchmark the performance of a model where we only sample $C=3$ random values, one for each channel, and then replicate them to fill the shape $W \times H$. In this case, the noise value is identical in each spatial location of a channel and only varies across channels. This model is referred to as \emph{Image-3C-Const}.

\begin{figure}
    \centering
    \includegraphics[width=\linewidth]{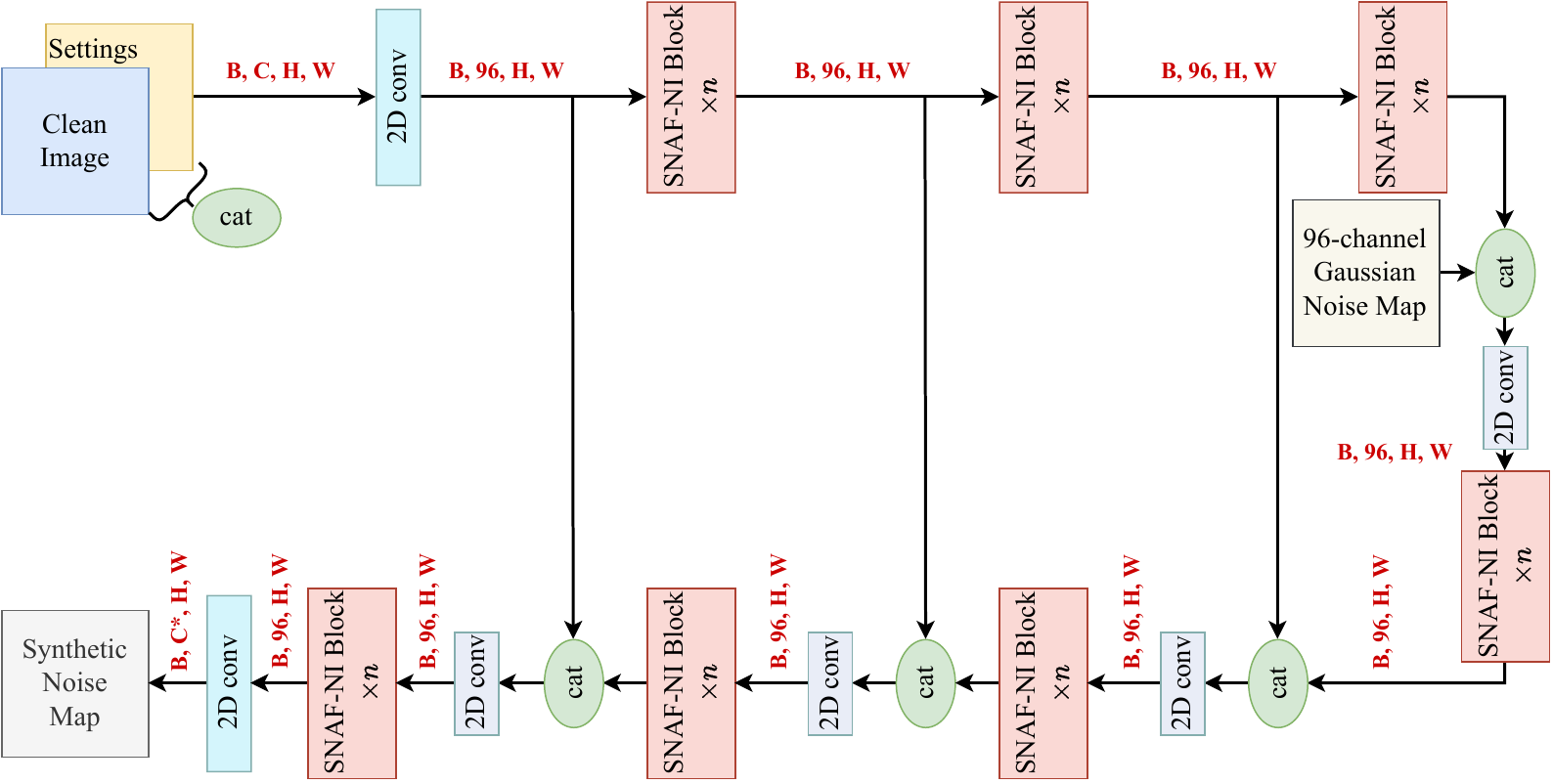}
    \caption{Ablation study on the position of noise injections, namely Full-NIN}
    \label{fig:ablation_2}
\end{figure}

\textbf{Importance of Clean Feature Guidance.} 
To show the importance of extracting features from the clean image as in our Clean Feature Guidance approach, we also compare to a variant where we use SNAF-NI blocks throughout the whole architecture. In this approach, noise is also injected in the earlier feature extraction stages of the network and we refer to this approach as \emph{Full-NIN}, as shown in Fig.~\ref{fig:ablation_2}.

By using the EM-1, as shown in Table~\ref{table:appendix_rgb_results}, the comparisons between \textit{Image-3C}, \textit{Image-96C}, and \textit{Latent-96C} indicated that conditioning on clean deep features is better than on clean image signals. We interpreted that concatenating the clean image, control map, and Gaussian noise together will negatively affect the conditioning mechanism since the intensity vs. variance curves in Fig. M4 in the main paper are irrelevant to the noise value. By using the EM-2, Table \ref{table:appendix_rgb_results_EM2} shows consistent performance comparison results. Furthermore, with the better dataset coverage and varieties of the testing set in the EM-2 (described in Sec. Experimental Setup in the main paper), the proposed CFG-NIN outperformed other variations, which verifies the effectiveness of the noise injection mechanism and the necessity of preserving the conditioned features clean.

\section{Evaluation Metrics}
\label{appendix:eval}
\subsection{EM-1}
\label{appendix:em_1}

As mentioned in Sec. 4.1 of the main paper, the metric utilized in EM-1 is sensitive to the spatial resolution of the patches. We test the same method with different patch sizes to demonstrate such behavior and provide the quantitative results on the sRGB noise modeling task in Table~\ref{table:res}. The bin width of the histogram is the same as the one described in the main paper. We use the same check-point of the network (CFG-NIN) as the one used in the main paper for quantitative comparison without any further training or fine-tuning throughout the appendices.

\begin{table}[]
\centering
\setlength\tabcolsep{3.2pt}
\begin{tabular}{bababab}
\hline\hline
Patch size & G4    & GP    & IP    & N6    & S6    & Agg   \\ \hline
32         & 0.026 & 0.015 & 0.012 & 0.037 & 0.025 & 0.021 \\
64         & 0.020 & 0.011 & 0.007 & 0.033 & 0.017 & 0.015 \\
128        & 0.016 & 0.008 & 0.005 & 0.029 & 0.014 & 0.012 \\
256        & 0.013 & 0.007 & 0.004 & 0.026 & 0.012 & 0.010 \\ \hline\hline
\end{tabular}
\caption{Quantitative results produced by CFG-NIN on sRGB noise modeling task given different patch sizes.}
\label{table:res}
\end{table}

\subsection{EM-2}
\label{appendix:em_2}

Different from the metric described in \cite{abdelhamed2019noise, maleky2022noise2noiseflow, kousha2022modeling}, \cite{jang2021c2n} adopted the evaluation method of the KL divergence from another perspective. Given a testing set containing multiple patches, ~\cite{jang2021c2n} stack all of them along the spatial dimension and compute the histogram on such a huge stacked image. To stay consistent with the Eq. (8) - (10) in the main paper, we rewrite such process which has been described in the supplementary material of \cite{jang2021c2n} as:
\begin{equation}
    bin_{n^*}(i)=\frac{1}{PCHW}\sum_{p,c,h,w}^{P,C,H,W}\mathbf{1}_{\{i\}}(n^*_{p,c,h,w}),
    \label{eq:em_2_0}
\end{equation}
where $p$ is the patch index, $c$ is the channel index, $h,w$ represent spatial position of a pixel in one patch, $n^*_{p,c,h,w}$ is the noise value at position $(p,c,h,w)$, and $i\in[-255, 255]$ is the possible noise value. Such a method utilizes the discrete nature of the pixel value in sRGB space without manually defining the range and bin width of the histogram. The calculation of KL divergence is similar to EM-1 and can be described as:
\begin{equation}
    D_{KL}^f=\sum_{i=-255}^{255}bin_{\tilde{n}}(i)\frac{bin_{\tilde{n}}(i)}{bin_{\hat{n}}(i)}.
    \label{eq:em_2_1}
\end{equation}

Although the above metric's non-parametric property is appealing, it disregards the spatial information and may result in error cancelation. However, it is still a meaningful metric for comparison between different methods. Moreover, while the metric in EM-1 reduces the spatial-wise error cancelation (which may arise in EM-2), EM-1 ignores the error cancelation happening between channels (see Appendix~\ref{appendix:quantitative} for more discussions). Hence we employed both EM-1 and EM-2 for complementing the evaluations from both aspects.

\section{Quantitative Results}
\label{appendix:quantitative}

We provide KL divergence results for each R, G and B channel estimated on the metric of EM-1 and EM-2 shown in Table~\ref{table:appendix_rgb_results} and Table~\ref{table:appendix_rgb_results_EM2} respectively. They are the same experiment as Table 1 and 2 in the main paper but show more details. As shown in Table~\ref{table:appendix_rgb_results}, the performance of \cite{kousha2022modeling} on B channel is comparably worse than R and G channels, which may be the source of the color shift shown in \cite{kousha2022modeling}'s visual results.

\begin{table*}[]
\small
\centering
\begin{tabular}{cc|ccc|cccccc}
\hline\hline
\multicolumn{2}{c|}{Method} & C2N\cite{jang2021c2n} & InvDN\cite{liu2021invertible} & \begin{tabular}[c]{@{}c@{}}sRGB\\ Flow~\cite{kousha2022modeling}\end{tabular} & \begin{tabular}[c]{@{}c@{}}Image-\\ 3C-Const\end{tabular} & Image-3C & Image-96C & Latent-96C & Full-NIN & CFG-NIN \\ \hline
\multicolumn{1}{c|}{} & R & 0.231 & 0.123 & 0.059 & 0.049 & 0.054 & 0.052 & 0.052 & 0.048 & 0.046 \\
\multicolumn{1}{c|}{} & G & 0.211 & 0.118 & 0.081 & 0.035 & 0.047 & 0.037 & 0.035 & 0.041 & 0.037 \\
\multicolumn{1}{c|}{} & B & 0.323 & 0.120 & 0.141 & 0.041 & 0.049 & 0.044 & 0.041 & 0.040 & 0.045 \\
\multicolumn{1}{c|}{\multirow{-4}{*}{G4}} & \cellcolor[HTML]{EFEFEF}Agg & \cellcolor[HTML]{EFEFEF}0.217 & \cellcolor[HTML]{EFEFEF}0.111 & \cellcolor[HTML]{EFEFEF}0.044 & \cellcolor[HTML]{EFEFEF}0.027 & \cellcolor[HTML]{EFEFEF}0.031 & \cellcolor[HTML]{EFEFEF}0.028 & \cellcolor[HTML]{EFEFEF}0.027 & \cellcolor[HTML]{EFEFEF}0.026 & \cellcolor[HTML]{EFEFEF}0.026 \\ \hline
\multicolumn{1}{c|}{} & R & 0.412 & 0.010 & 0.091 & 0.029 & 0.028 & 0.030 & 0.027 & 0.028 & 0.026 \\
\multicolumn{1}{c|}{} & G & 0.414 & 0.010 & 0.068 & 0.022 & 0.020 & 0.023 & 0.020 & 0.021 & 0.021 \\
\multicolumn{1}{c|}{} & B & 0.463 & 0.012 & 0.103 & 0.029 & 0.027 & 0.029 & 0.030 & 0.028 & 0.027 \\
\multicolumn{1}{c|}{\multirow{-4}{*}{GP}} & \cellcolor[HTML]{EFEFEF}Agg & \cellcolor[HTML]{EFEFEF}0.426 & \cellcolor[HTML]{EFEFEF}0.008 & \cellcolor[HTML]{EFEFEF}0.059 & \cellcolor[HTML]{EFEFEF}0.017 & \cellcolor[HTML]{EFEFEF}0.016 & \cellcolor[HTML]{EFEFEF}0.017 & \cellcolor[HTML]{EFEFEF}0.016 & \cellcolor[HTML]{EFEFEF}0.016 & \cellcolor[HTML]{EFEFEF}0.015 \\ \hline
\multicolumn{1}{c|}{} & R & 0.130 & 0.076 & 0.035 & 0.031 & 0.039 & 0.041 & 0.034 & 0.033 & 0.032 \\
\multicolumn{1}{c|}{} & G & 0.114 & 0.087 & 0.038 & 0.021 & 0.021 & 0.021 & 0.018 & 0.020 & 0.017 \\
\multicolumn{1}{c|}{} & B & 0.185 & 0.081 & 0.066 & 0.029 & 0.030 & 0.029 & 0.032 & 0.027 & 0.025 \\
\multicolumn{1}{c|}{\multirow{-4}{*}{IP}} & \cellcolor[HTML]{EFEFEF}Agg & \cellcolor[HTML]{EFEFEF}0.114 & \cellcolor[HTML]{EFEFEF}0.077 & \cellcolor[HTML]{EFEFEF}0.020 & \cellcolor[HTML]{EFEFEF}0.013 & \cellcolor[HTML]{EFEFEF}0.014 & \cellcolor[HTML]{EFEFEF}0.015 & \cellcolor[HTML]{EFEFEF}0.013 & \cellcolor[HTML]{EFEFEF}0.012 & \cellcolor[HTML]{EFEFEF}0.012 \\ \hline
\multicolumn{1}{c|}{} & R & 0.244 & 0.053 & 0.095 & 0.058 & 0.067 & 0.079 & 0.058 & 0.056 & 0.060 \\
\multicolumn{1}{c|}{} & G & 0.223 & 0.062 & 0.081 & 0.034 & 0.037 & 0.044 & 0.034 & 0.034 & 0.035 \\
\multicolumn{1}{c|}{} & B & 0.300 & 0.054 & 0.116 & 0.052 & 0.061 & 0.057 & 0.052 & 0.048 & 0.051 \\
\multicolumn{1}{c|}{\multirow{-4}{*}{N6}} & \cellcolor[HTML]{EFEFEF}Agg & \cellcolor[HTML]{EFEFEF}0.230 & \cellcolor[HTML]{EFEFEF}0.051 & \cellcolor[HTML]{EFEFEF}0.062 & \cellcolor[HTML]{EFEFEF}0.034 & \cellcolor[HTML]{EFEFEF}0.041 & \cellcolor[HTML]{EFEFEF}0.045 & \cellcolor[HTML]{EFEFEF}0.034 & \cellcolor[HTML]{EFEFEF}0.032 & \cellcolor[HTML]{EFEFEF}0.037 \\ \hline
\multicolumn{1}{c|}{} & R & 0.436 & 0.227 & 0.066 & 0.043 & 0.048 & 0.058 & 0.041 & 0.044 & 0.043 \\
\multicolumn{1}{c|}{} & G & 0.493 & 0.143 & 0.096 & 0.037 & 0.045 & 0.049 & 0.036 & 0.039 & 0.039 \\
\multicolumn{1}{c|}{} & B & 0.624 & 0.143 & 0.147 & 0.046 & 0.054 & 0.062 & 0.045 & 0.049 & 0.048 \\
\multicolumn{1}{c|}{\multirow{-4}{*}{S6}} & \cellcolor[HTML]{EFEFEF}Agg & \cellcolor[HTML]{EFEFEF}0.541 & \cellcolor[HTML]{EFEFEF}0.161 & \cellcolor[HTML]{EFEFEF}0.050 & \cellcolor[HTML]{EFEFEF}0.023 & \cellcolor[HTML]{EFEFEF}0.029 & \cellcolor[HTML]{EFEFEF}0.036 & \cellcolor[HTML]{EFEFEF}0.023 & \cellcolor[HTML]{EFEFEF}0.024 & \cellcolor[HTML]{EFEFEF}0.025 \\ \hline
\multicolumn{2}{c|}{Agg} & 0.335 & 0.092 & 0.044 & 0.021 & 0.024 & 0.027 & 0.020 & 0.020 & 0.021 \\ \hline\hline
\end{tabular}
\caption{KL divergence results for each R, G and B channel on sRGB noise modeling task by using the method in EM-1.}
\label{table:appendix_rgb_results}
\end{table*}

\begin{table*}
\small
\centering
\begin{tabular}{cc|cc|cccccc}
\hline\hline
\multicolumn{2}{c|}{Method} & C2N~\cite{jang2021c2n} & InvDN~\cite{liu2021invertible} & \begin{tabular}[c]{@{}c@{}}Image-\\ 3C-Const\end{tabular} & Image-3C & Image-96C & Latent-96C & Full-NIN & CFG-NIN \\ \hline
\multicolumn{1}{c|}{} & R & 11.55 & 3.823 & 0.245 & 0.230 & 0.164 & 0.117 & 0.112 & 0.128 \\
\multicolumn{1}{c|}{} & G & 10.80 & 4.116 & 0.155 & 0.146 & 0.261 & 0.128 & 0.204 & 0.092 \\
\multicolumn{1}{c|}{} & B & 14.00 & 3.464 & 0.189 & 0.091 & 0.104 & 0.071 & 0.122 & 0.108 \\
\multicolumn{1}{c|}{\multirow{-4}{*}{G4}} & \cellcolor[HTML]{EFEFEF}Agg & \cellcolor[HTML]{EFEFEF}8.970 & \cellcolor[HTML]{EFEFEF}3.360 & \cellcolor[HTML]{EFEFEF}0.154 & \cellcolor[HTML]{EFEFEF}0.105 & \cellcolor[HTML]{EFEFEF}0.093 & \cellcolor[HTML]{EFEFEF}0.072 & \cellcolor[HTML]{EFEFEF}0.083 & \cellcolor[HTML]{EFEFEF}0.071 \\ \hline
\multicolumn{1}{c|}{} & R & 18.48 & 2.112 & 0.122 & 0.151 & 0.107 & 0.070 & 0.080 & 0.071 \\
\multicolumn{1}{c|}{} & G & 20.10 & 1.894 & 0.053 & 0.089 & 0.306 & 0.123 & 0.149 & 0.153 \\
\multicolumn{1}{c|}{} & B & 28.25 & 1.966 & 0.116 & 0.104 & 0.163 & 0.098 & 0.063 & 0.051 \\
\multicolumn{1}{c|}{\multirow{-4}{*}{GP}} & \cellcolor[HTML]{EFEFEF}Agg & \cellcolor[HTML]{EFEFEF}18.54 & \cellcolor[HTML]{EFEFEF}1.793 & \cellcolor[HTML]{EFEFEF}0.068 & \cellcolor[HTML]{EFEFEF}0.087 & \cellcolor[HTML]{EFEFEF}0.112 & \cellcolor[HTML]{EFEFEF}0.066 & \cellcolor[HTML]{EFEFEF}0.057 & \cellcolor[HTML]{EFEFEF}0.067 \\ \hline
\multicolumn{1}{c|}{} & R & 3.472 & 5.172 & 0.278 & 0.234 & 0.248 & 0.131 & 0.144 & 0.127 \\
\multicolumn{1}{c|}{} & G & 4.126 & 6.635 & 0.215 & 0.122 & 0.153 & 0.124 & 0.366 & 0.164 \\
\multicolumn{1}{c|}{} & B & 10.24 & 6.967 & 0.237 & 0.214 & 0.331 & 0.200 & 0.288 & 0.206 \\
\multicolumn{1}{c|}{\multirow{-4}{*}{IP}} & \cellcolor[HTML]{EFEFEF}Agg & \cellcolor[HTML]{EFEFEF}3.070 & \cellcolor[HTML]{EFEFEF}5.483 & \cellcolor[HTML]{EFEFEF}0.176 & \cellcolor[HTML]{EFEFEF}0.149 & \cellcolor[HTML]{EFEFEF}0.130 & \cellcolor[HTML]{EFEFEF}0.099 & \cellcolor[HTML]{EFEFEF}0.145 & \cellcolor[HTML]{EFEFEF}0.106 \\ \hline
\multicolumn{1}{c|}{} & R & 41.99 & 9.331 & 0.119 & 0.226 & 0.163 & 0.065 & 0.113 & 0.085 \\
\multicolumn{1}{c|}{} & G & 48.97 & 4.344 & 0.248 & 0.165 & 0.251 & 0.152 & 0.420 & 0.142 \\
\multicolumn{1}{c|}{} & B & 55.17 & 5.861 & 0.148 & 0.121 & 0.179 & 0.134 & 0.100 & 0.078 \\
\multicolumn{1}{c|}{\multirow{-4}{*}{N6}} & \cellcolor[HTML]{EFEFEF}Agg & \cellcolor[HTML]{EFEFEF}38.69 & \cellcolor[HTML]{EFEFEF}5.806 & \cellcolor[HTML]{EFEFEF}0.094 & \cellcolor[HTML]{EFEFEF}0.118 & \cellcolor[HTML]{EFEFEF}0.102 & \cellcolor[HTML]{EFEFEF}0.061 & \cellcolor[HTML]{EFEFEF}0.076 & \cellcolor[HTML]{EFEFEF}0.047 \\ \hline
\multicolumn{1}{c|}{} & R & 44.81 & 7.820 & 0.231 & 0.327 & 0.355 & 0.275 & 0.283 & 0.283 \\
\multicolumn{1}{c|}{} & G & 52.38 & 3.907 & 0.170 & 0.248 & 0.350 & 0.243 & 0.389 & 0.161 \\
\multicolumn{1}{c|}{} & B & 63.02 & 5.287 & 0.130 & 0.206 & 0.240 & 0.174 & 0.156 & 0.153 \\
\multicolumn{1}{c|}{\multirow{-4}{*}{S6}} & \cellcolor[HTML]{EFEFEF}Agg & \cellcolor[HTML]{EFEFEF}45.29 & \cellcolor[HTML]{EFEFEF}5.261 & \cellcolor[HTML]{EFEFEF}0.120 & \cellcolor[HTML]{EFEFEF}0.184 & \cellcolor[HTML]{EFEFEF}0.208 & \cellcolor[HTML]{EFEFEF}0.156 & \cellcolor[HTML]{EFEFEF}0.161 & \cellcolor[HTML]{EFEFEF}0.151 \\ \hline
\multicolumn{2}{c|}{Agg} & 14.06 & 5.204 & 0.048 & 0.065 & 0.071 & \blue 0.062 \black & 0.063 & \red 0.055 \black \\ \hline\hline
\end{tabular}
\caption{KL divergence results for each R, G and B channel on sRGB noise modeling task by using the method in EM-2. The results are multiplied by $10^2$ for better readability. The best and second best methods are in \red red \black and \blue blue \black, respectively. Due to the dragging artifacts, Image-3C-Const is excluded from the ranking.}
\label{table:appendix_rgb_results_EM2}
\end{table*}

\section{Visual Results}
\label{appendix:visual}

As mentioned in Sec.~\ref{subsec:visual} of the main paper, we provide the visual results of CFG-NIN trained with the dataset splitting method described in EM-2. As shown in Fig.~\ref{fig:srgb_visual_appendix}-\ref{fig:raw_visual_appendix}, CFG-NIN can generate noise with the correct pattern and magnitude on relatively high ISO levels.

To visualize the temporal variance and dragging artifacts mentioned in Sec.~\ref{subsec:temporal_variance} of the main paper intuitively, we provide \textbf{videos} in the folder \textbf{\textit{temporal\_variance}} for better comparison. We synthesize noise 150 times with different random seeds, and use the noisy images in SIDD Full~\cite{abdelhamed2018high} as ground truth. We increase the brightness of the noisy images and noise maps for easier visualization and comparison.

\section{Denoising Performance}
\label{appendix:denoising}
\begin{figure}
    \centering
    \begin{subfigure}{\linewidth}
        \includegraphics[width=\linewidth]{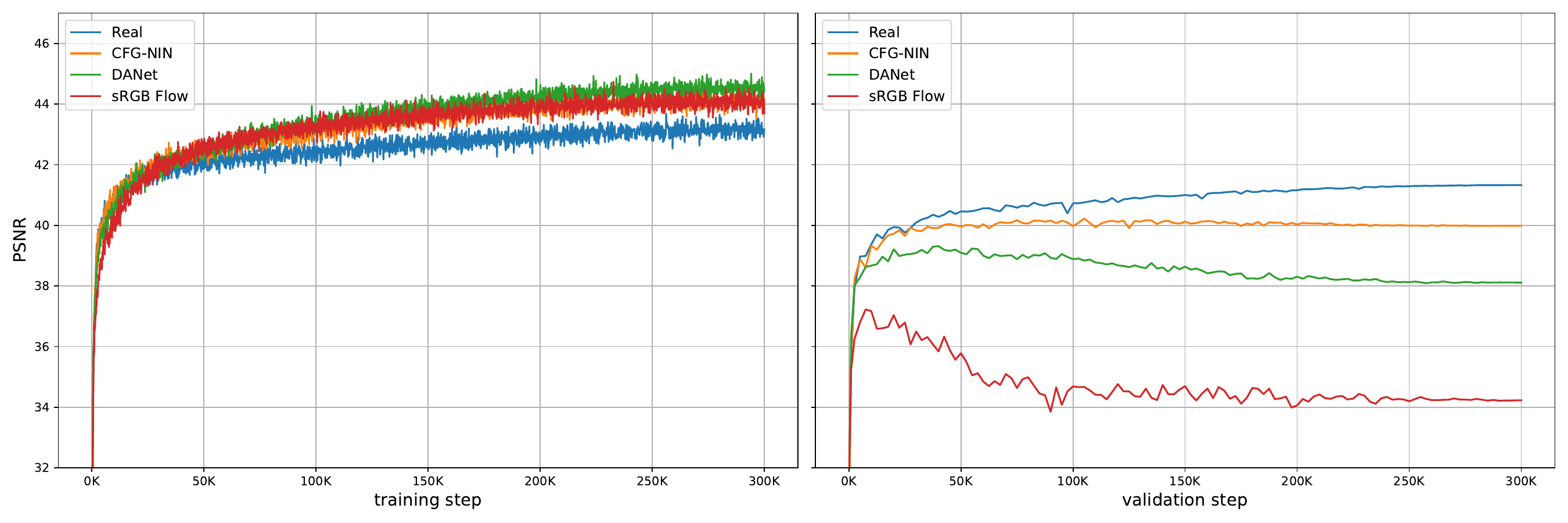}
        \caption{NAF~\cite{chen2022simple}}
        \label{fig:naf_srgb}
    \end{subfigure}
    \hfill
    \begin{subfigure}{\linewidth}
        \includegraphics[width=\linewidth]{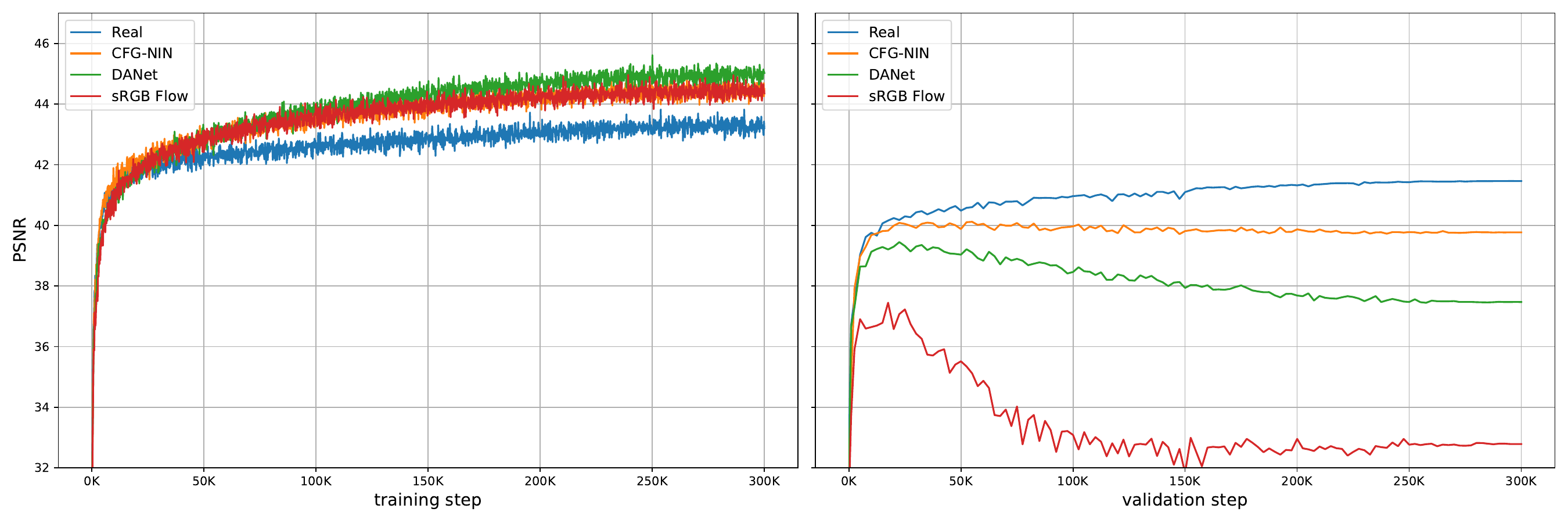}
        \caption{Restormer~\cite{zamir2022restormer}}
        \label{fig:restormer_srgb}
    \end{subfigure}
    \caption{PSNR curves of the denoisers trained with real noise and different synthetic noise in sRGB space. Left: PSNR vs. training step. Right: PSNR vs. testing step.}
    \label{fig:psnr_gap_srgb}
\end{figure}

\begin{figure}
    \centering
    \begin{subfigure}{\linewidth}
        \includegraphics[width=\linewidth]{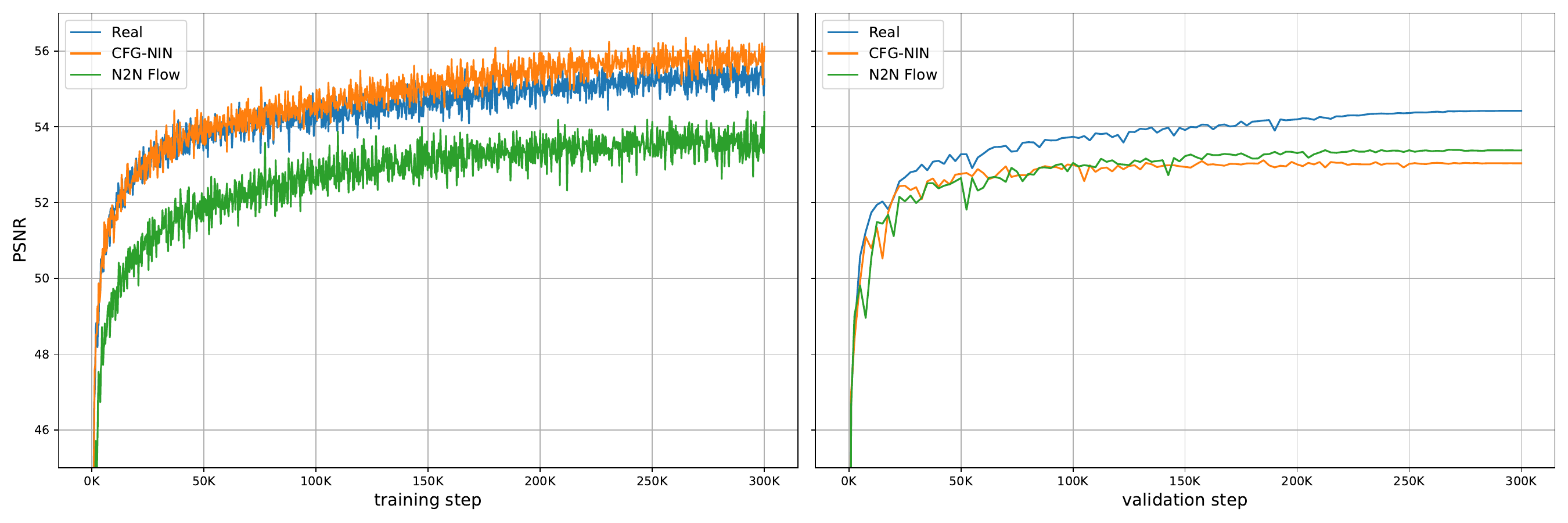}
        \caption{NAF~\cite{chen2022simple}}
        \label{fig:naf_raw}
    \end{subfigure}
    \hfill
    \begin{subfigure}{\linewidth}
        \includegraphics[width=\linewidth]{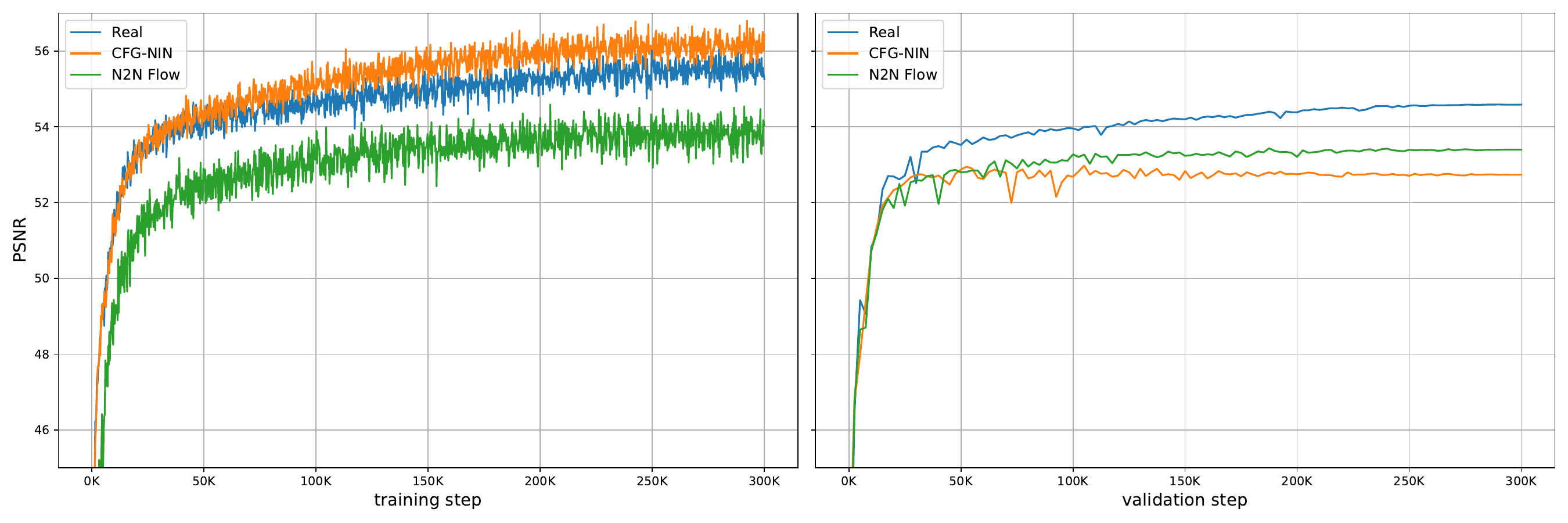}
        \caption{Restormer\cite{zamir2022restormer}}
        \label{fig:restormer_raw}
    \end{subfigure}
    \caption{PSNR curves of the denoisers trained with real noise and different synthetic noise in rawRGB space. Left: PSNR vs. training step. Right: PSNR vs. testing step.}
    \label{fig:psnr_gap_raw}
\end{figure}

\begin{figure*}[]
    \centering
    \includegraphics[width=0.9\linewidth]{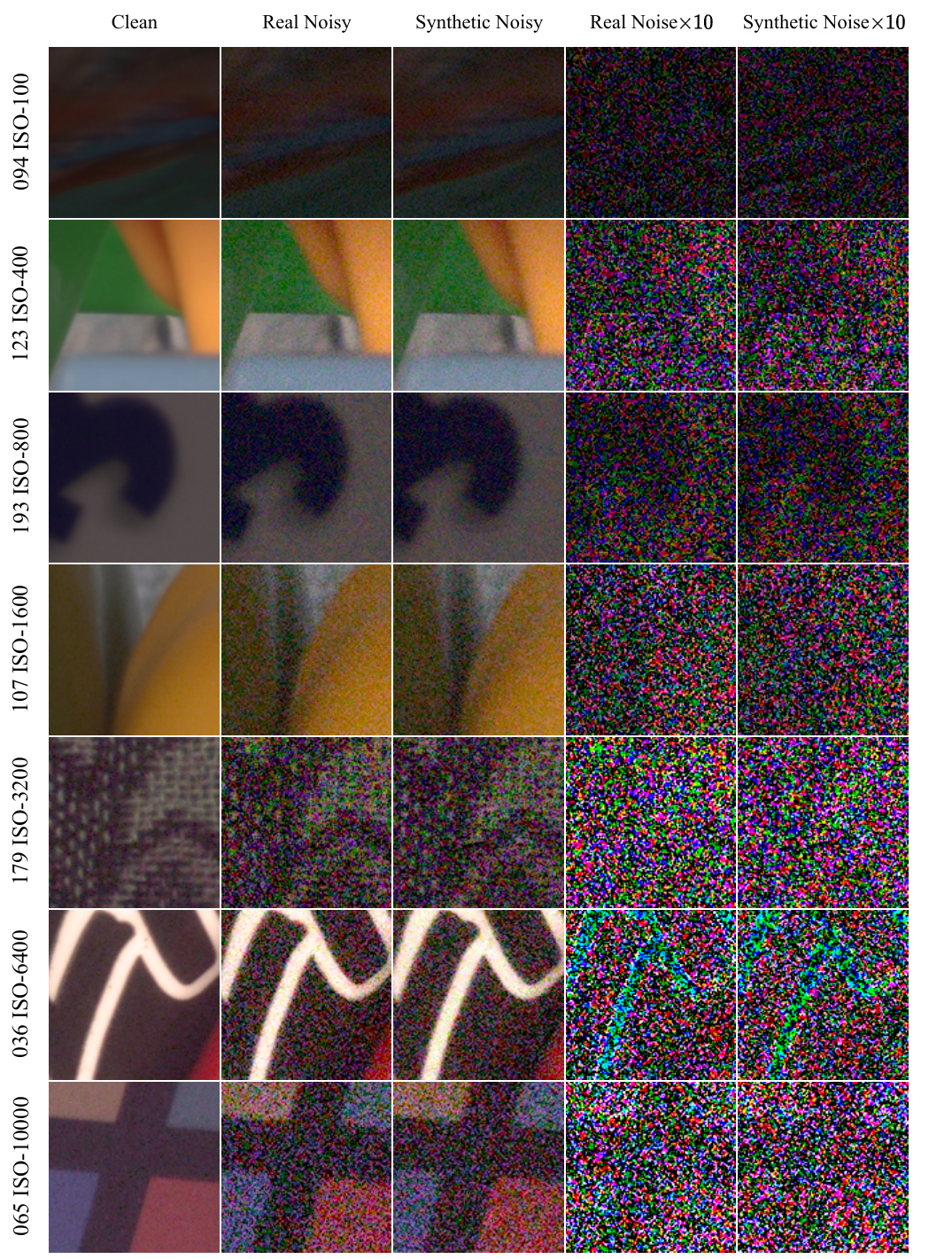}
    \caption{The synthesized noisy images and noise maps by CFG-NIN trained according to EM-2 on the sRGB noise modeling task.}
    \label{fig:srgb_visual_appendix}
\end{figure*}

\begin{figure*}[]
    \centering
    \includegraphics[width=0.9\linewidth]{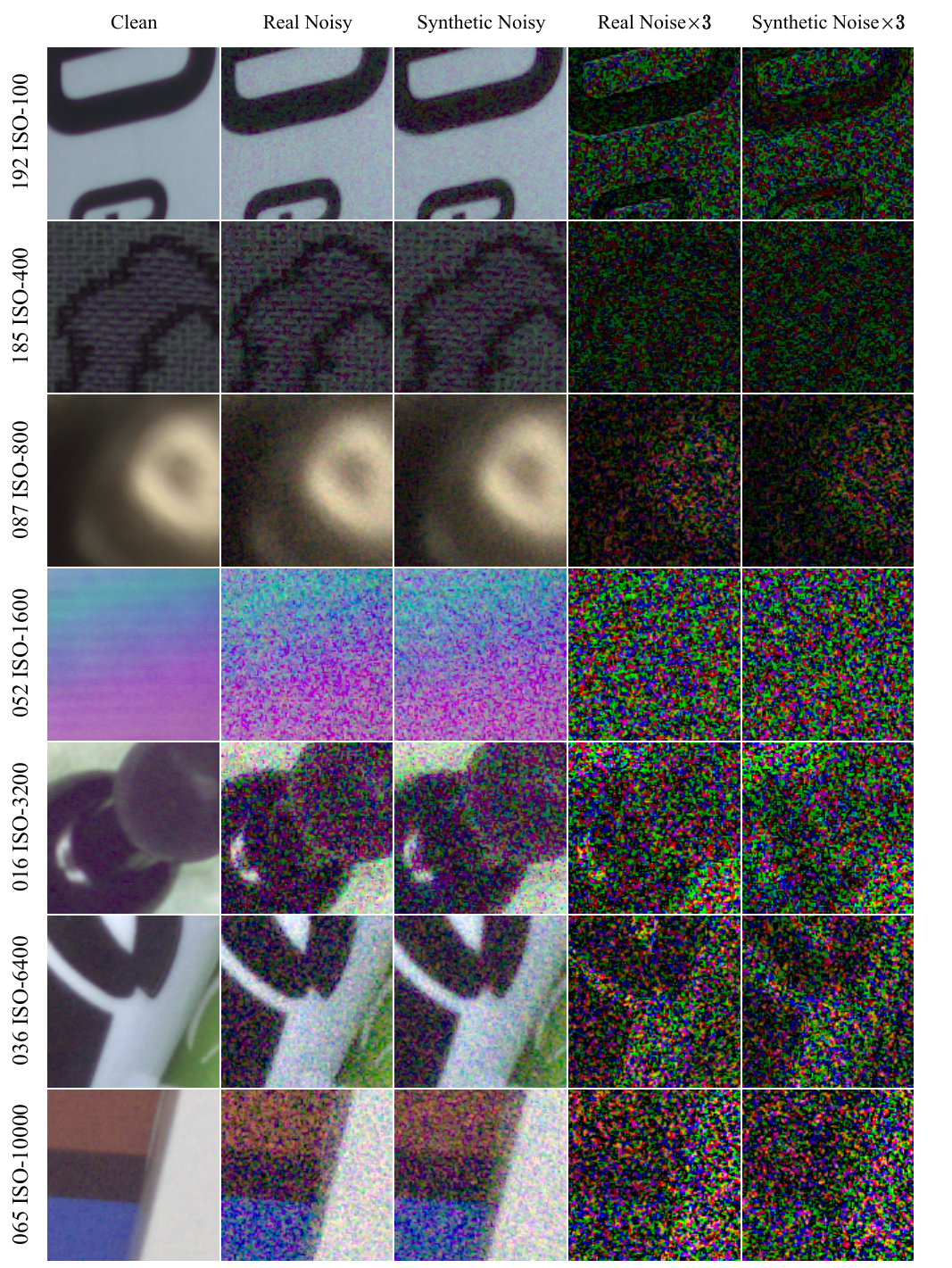}
    \caption{The synthesized noisy images and noise maps by CFG-NIN trained according to EM-2 on the rawRGB noise modeling task.}
    \label{fig:raw_visual_appendix}
\end{figure*}

This paper mainly focuses on the visual results of the synthesized noisy images/videos, which were thoroughly evaluated by $D_{KL}$, temporal variance, and spatial correlation in the main paper. The experiments of denoising provide an indirect measurement of the similarity between real and synthetic noise. As mentioned in Sec.~\ref{subsec:quantitative_results} in the main paper, we use two recent SOTA image restoration models, NAF~\cite{chen2022simple} and Restormer~\cite{zamir2022restormer} as the denoisers. The training curves on the left-hand side of Fig.~\ref{fig:psnr_gap_srgb}-~\ref{fig:psnr_gap_raw} represent how well the denoisers fit the distribution of synthetic noise. The testing curves on the right represent the denoising performance on real noisy images. We use the denoisers trained with real data as the reference and plot its training and testing curves together with the synthetic ones. The PSNR Gap~\cite{yue2020dual} is described as the difference between the performance of the denoisers trained on the real and synthetic data when denoising the real noisy images, which is visually demonstrated in the right-hand side of Fig.~\ref{fig:psnr_gap_srgb}-~\ref{fig:psnr_gap_raw}.

We provide the quantitative PSNR Gaps in Table~\ref{table:psnr_gap_srgb}-\ref{table:psnr_gap_raw}. As shown in Table~\ref{table:psnr_gap_srgb}, our models achieved minimum PSNR Gaps on both denoisers in sRGB space, indicating the distribution generated by CFG-NIN is closer to the real one. In rawRGB space, our models show comparable performance with SOTA method N2N Flow~\cite{maleky2022noise2noiseflow}, as shown in Table~\ref{table:psnr_gap_raw}.

To keep consistent with Table 4 in the main paper, we provide the denoising performance of NAF~\cite{chen2022simple} and Restormer~\cite{zamir2022restormer} on the SIDD benchmark. As shown in Table~\ref{table:bench_srgb}, the denoisers trained with CFG-NIN synthetic noise achieved better performance than the other two synthetic methods. Note that the SIDD benchmark contains several high ISO contents whose noise distribution is out of the training set of sRGB Flow~\cite{kousha2022modeling}, which is mentioned in EM-1 in the main paper. As a result, the performance of the denoiser trained with sRGB Flow~\cite{kousha2022modeling} synthetic noise is affected by the visible color shift under high ISO. As shown in Table~\ref{table:bench_raw}, benefiting from the larger training set (SIDD Full), the performance of denoisers trained with N2N Flow~\cite{maleky2022noise2noiseflow} synthetic noise is closer to real one, which is consistent with the experiments of PSNR Gaps.

\begin{table}[]
    \small
    \centering
    \begin{tabular}{ccc}
    \hline \hline
    \begin{tabular}[c]{@{}c@{}}Denoiser\\ (PSNR$\uparrow$/PSNR Gap$\downarrow$)\end{tabular} & NAF~\cite{chen2022simple} & Restormer~\cite{zamir2022restormer} \\ \hline 
    sRGB Flow~\cite{kousha2022modeling} & 33.96/6.93 & 32.55/8.47 \\
    DANet~\cite{yue2020dual} & \blue 37.96/2.93 \black & \blue 37.32/3.70 \black \\
    CFG-NIN$^{\ssymbol{4}}$ & \red 39.67/1.22 \black & \red 39.44/1.58 \black \\ \hline
    Real & 40.89/-- & 41.02/--  \\ \hline \hline
    \end{tabular}
    \caption{PSNR and PANR Gaps of different noise modeling models in sRGB space. ${\ssymbol{4}}$: the synthetic noise is generated by the CFG-NIN trained on EM-2.}
    \label{table:psnr_gap_srgb}
\end{table}

\begin{table}[]
    \small
    \centering
    \begin{tabular}{ccc}
    \hline \hline
    \begin{tabular}[c]{@{}c@{}}Denoiser\\ (PSNR$\uparrow$/PSNR Gap$\downarrow$)\end{tabular} & NAF~\cite{chen2022simple} & Restormer~\cite{zamir2022restormer} \\ \hline
    N2N Flow~\cite{maleky2022noise2noiseflow} & \red 52.00/1.15 \black & \red 51.98/1.37 \black \\
    CFG-NIN$^{\ssymbol{4}}$ & \blue 51.75/1.40 \black & \blue 51.47/1.88 \black \\ \hline
    Real & 53.15/-- & 53.35/-- \\ \hline \hline
    \end{tabular}
    \caption{PSNR and PANR Gaps of different noise modeling models in rawRGB space. ${\ssymbol{4}}$: the synthetic noise is generated by the CFG-NIN trained on EM-2.}
    \label{table:psnr_gap_raw}
\end{table}

\begin{table}[]
    \small
    \centering
    \begin{tabular}{ccc}
    \hline \hline
    Denoiser(PSNR$\uparrow$/SSIM$\uparrow$) & NAF~\cite{chen2022simple} & Restormer~\cite{zamir2022restormer} \\ \hline
    sRGB Flow~\cite{kousha2022modeling} & 30.20/0.809 & 28.77/0.772 \\
    DANet~\cite{yue2020dual} & \blue 34.52/0.910 \black & \blue 34.04/0.901 \black \\
    CFG-NIN$^{\ssymbol{4}}$ & \red 36.20/0.929 \black & \red 36.13/0.927 \black \\ \hline
    Real & 37.31/0.943 & 37.33/0.943 \\ \hline \hline
    \end{tabular}
    \caption{Comparison of denoising performance (PSNR and SSIM) between denoisers trained on real and synthetic noise in sRGB space. ${\ssymbol{4}}$: the synthetic noise is generated by the CFG-NIN trained on EM-2.}
    \label{table:bench_srgb}
\end{table}

\begin{table}[]
    \small
    \centering
    \begin{tabular}{ccc}
    \hline \hline
    Denoiser(PSNR$\uparrow$/SSIM$\uparrow$) & NAF~\cite{chen2022simple} & Restormer~\cite{zamir2022restormer} \\ \hline
    N2N Flow~\cite{maleky2022noise2noiseflow} & \red 48.32/0.984 \black & \red 48.35/0.984 \black \\
    CFG-NIN$^{\ssymbol{4}}$ & \blue 47.70/0.983 \black & \blue 47.47/0.983 \black \\ \hline
    Real & 48.50/0.985 & 48.55/0.985 \\ \hline \hline
    \end{tabular}
    \caption{Comparison of denoising performance (PSNR and SSIM) between denoisers trained on real and synthetic noise in rawRGB space. ${\ssymbol{4}}$: the synthetic noise is generated by the CFG-NIN trained on EM-2.}
    \label{table:bench_raw}
\end{table}

\section{Inference Details}
To synthesize noise on images with arbitrary spatial resolution and keep the optimal performance of CFG-NIN, we tiled the whole image into multiple overlapping $128\times128$ patches. The overlap margin is set to $25$ pixels, which can be adjusted according to the practical use case.

\section{Sampling Strategy in Temporal Variance Experiments}
\label{appendix:sample_remporal}
Temporal variance is a crucial aspect of the noise synthesis tasks, especially for video renoising use cases. As mentioned in Sec.\ref{subsec:temporal_variance} in the main paper, we describe the strategy of collecting a large number of noise samples when the number of noisy images is limited. For the ground truth noise samples (labeled \textit{Real} in Fig. 8 and 9 in the main paper), we first crop one flat patch, where the clean intensity values are almost the same, to ensure the noise values on it have similar conditions. Such a patch can be extracted from a color chart scene (001) in SIDD Full~\cite{abdelhamed2018high}. For each clean image, SIDD Full~\cite{abdelhamed2018high} provides 150 pairing noise images, which means for a specific pixel position, there are 150 corresponding noise samples. To enlarge the number of samples, considering the spatial correlation, we choose pixels on the patch with a stride of 10 to ensure the noise values are independent (as shown in Fig. 10 in the main paper that noise shows independent characteristics with a stride larger than 2). As a result, we finally collect $N\times150$ noise values, where $N$ is the number of sampled positions on the patch.

\end{appendices}

\end{document}